\documentclass[runningheads]{llncs}

\usepackage{accv}

\usepackage{accvabbrv}
\usepackage{graphicx}
\usepackage{booktabs}
\usepackage{multirow}
\usepackage{array}
\usepackage{bm}
\usepackage{wrapfig}
\usepackage{needspace}
\usepackage{amssymb}
\usepackage{listings}

\definecolor{fix}{rgb}{0,0,0}

\IfFileExists{ulem.sty}{%
  \usepackage[normalem]{ulem}%
}{%
  \newcommand{\uline}[1]{\underline{#1}}%
}

\IfFileExists{fontawesome5.sty}{%
  \usepackage{fontawesome5}%
}{%
  \newcommand{\faSnowflake}{\ensuremath{\ast}}%
  \newcommand{\faFire}{\ensuremath{\blacktriangle}}%
}

\IfFileExists{axessibility.sty}{%
  \usepackage[accsupp]{axessibility}%
}{}

\usepackage[breaklinks,colorlinks,citecolor=accvblue]{hyperref}

\usepackage{orcidlink}
\renewcommand{\orcidID}[1]{\unskip\,\orcidlink{#1}}

\newcommand{\PaperTitle}{STARE: Scene Text-Aware multimodal Retrieval for Everyday objects}
\newcommand{\PaperTitleRunning}{STARE: Scene Text-Aware multimodal Retrieval}

\newcommand{\PaperAuthors}{%
  Kento Tokura\orcidID{0009-0009-4524-9022} \and
  Ryosuke Korekata\orcidID{0009-0008-0354-9070} \and
  Daichi Yashima\orcidID{0009-0005-2087-2038} \and
  Kei Katsumata\orcidID{0009-0009-7078-457X} \and
  Kanon Amemiya\orcidID{0009-0001-9655-7561} \and
  Miyu Goko\orcidID{0009-0009-9233-7121} \and
  Komei Sugiura\orcidID{0000-0002-0261-0510}}
\newcommand{\PaperAuthorsRunning}{K.~Tokura et al.}
\newcommand{\PaperInstitute}{%
  Keio University, Japan\\
  \email{\{tkento1985, rkorekata, ydaichi1207, ke59ka77, kanon-amemiya, miyu.goko, komei.sugiura\}@keio.jp}}

\begin{document}

\title{\PaperTitle}

\titlerunning{\PaperTitleRunning}

\author{\PaperAuthors}
\authorrunning{\PaperAuthorsRunning}

\institute{\PaperInstitute}

\maketitle

\begin{abstract}
In this study, we focus on service robots that retrieve and transport everyday objects based on free-form instructions that may refer to visual features and scene text.
This is challenging because it requires a multimodal understanding of the visual features and spatial relationships of objects, as well as scene text embedded in the environment.
To address this, we propose Scene Text-Aware multimodal Retrieval for Everyday objects (STARE), which leverages visual and textual cues in response to free-form instructions.
STARE introduces Crosslingual Visual Prompt to enhance scene text interpretation, and models the complex relationships between named entities in the instructions and their corresponding objects.
We evaluate STARE using two novel benchmarks, GoGetIt and TextCaps-test, both of which consist of real-world indoor/outdoor images and instructions involving object manipulation and navigation.
The results show that STARE outperforms all baseline methods on standard multimodal retrieval metrics.
Furthermore, in physical experiments conducted under a zero-shot transfer setting, STARE achieves an overall success rate of 80\%. 
Our project page is available at \url{https://stare-projectpage.pages.dev/}.
 \end{abstract}

\section{Introduction}

Service robots in indoor and outdoor environments such as households, retail stores, and urban streets are becoming increasingly important due to labor shortages and aging populations.
When interacting with service robots, 
it is natural for users to refer to objects using 
scene text
(e.g., labels, signage, packaging), which provides a
clearer and more direct way of 
specifying
objects than relying solely on visual features.
Therefore, in real-world applications,
it is crucial that robots can
retrieve target objects
based on
user instructions
by leveraging
scene text and visual information.

{
In this study, we formulate this problem as Scene Text-Aware Multimodal Retrieval (STMR).
Given a free-form natural language instruction for object manipulation or navigation and a set of images collected during pre-exploration, STMR retrieves and ranks images containing the target object.
}
The task considers both images with and without scene text.

\begin{figure}[t]
    \centering
    \includegraphics[clip,width=0.8\linewidth]{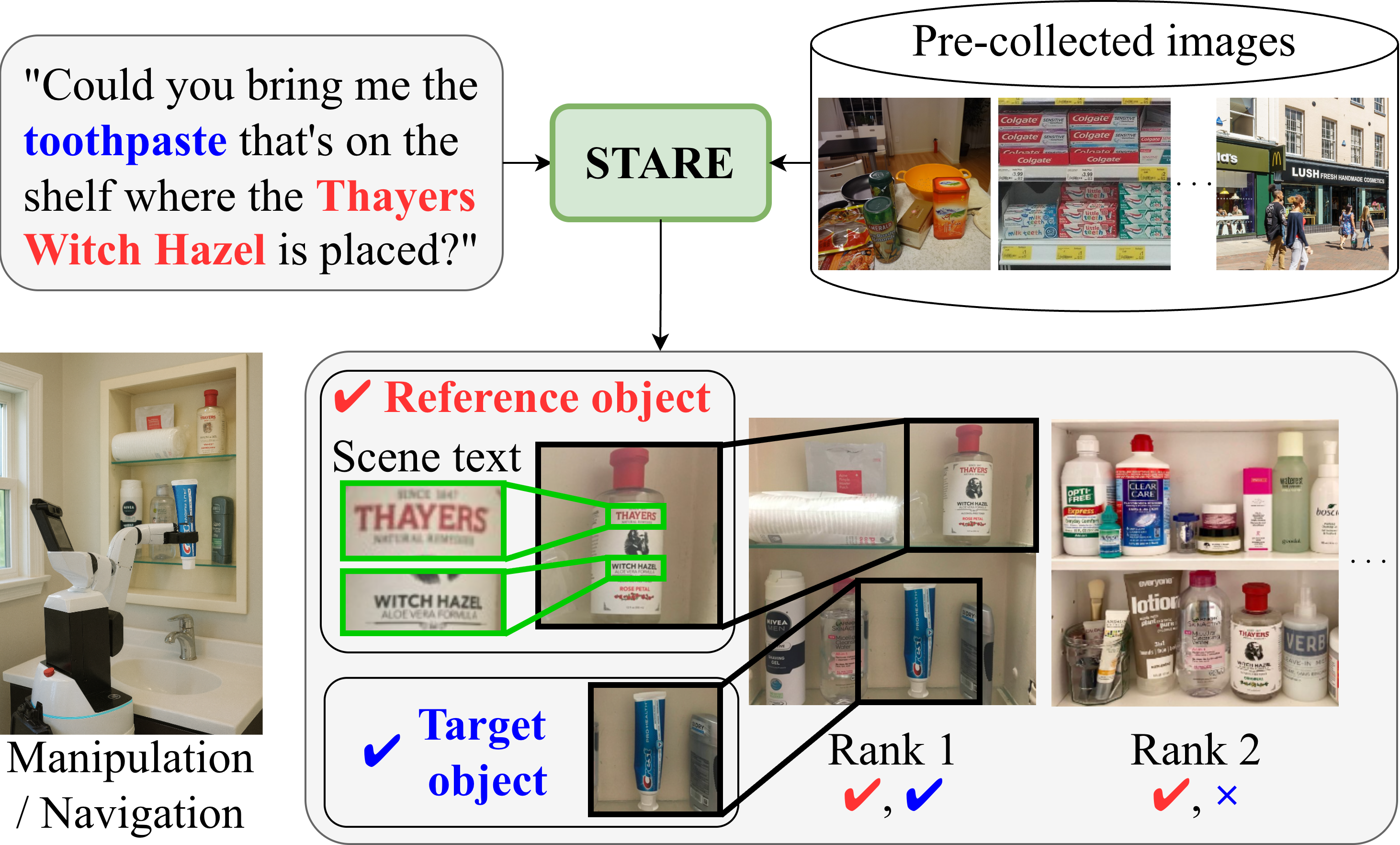}
    \caption{
    \textbf{Typical scene of our task}. 
    First, environmental images are collected in a pre-exploration phase.
    Next, STARE performs multimodal retrieval in response to a free-form instruction.
    Finally, the robot navigates to and grasps the specified object.
    }
    \label{fig:eyecatch}
    \vspace{-2mm}
\end{figure}
Fig.~\ref{fig:eyecatch} shows a typical scene of the task.
In the example, a user gives the instruction: ``Could you bring me the toothpaste that's on the shelf where the Thayers Witch Hazel is placed?''
The robot then searches through previously collected images for a toothpaste tube located on the same shelf as a bottle labeled “Thayers Witch Hazel” and performs the specified manipulation.

This task is challenging because it requires an understanding of not only the visual features and spatial relationships among objects but also scene text.
While existing methods \cite{clip, beit3} have been successfully applied to multimodal retrieval, their integration of scene text with other multimodal information is not always sufficient. 
Linking named entities in instructions to scene text associated with the corresponding objects remains challenging, limiting performance on this task.
Furthermore, because multimodal retrieval requires large-scale and fast inference, relying solely on multimodal large language models (MLLMs) is not practical.

To address these limitations, we propose Scene Text-Aware multimodal Retrieval for Everyday objects (STARE) for the STMR task.
STARE consists of three modules: a Multi-Form Instruction Encoder (MFIE) for encoding free-form instructions, a Scene Text-aware Visual Encoder (STVE) for encoding candidate images, and a Scene Text ReRanker (STRR) for reranking candidates based on scene-text cues.
Within STVE, we introduce Crosslingual Visual Prompt (CVP), which visually associates scene text detected by optical character recognition (OCR) with its image location while reducing confusion between the prompt marks and the original scene text.
The pre-exploration can be performed up
until query time to dynamically update the collected images.
Our contributions can be stated as follows.\nopagebreak[4]
\begin{itemize}
    \setlength{\parskip}{0.2mm} 
    \setlength{\itemsep}{0.2mm} 
    \item STVE, which 
    combines
    image features from encoders, fine-grained image features using overlapped patchification, and narrative representations that consider scene text.
    \item We introduce the STRR module, which efficiently models the highly complex relationships between named entities and their corresponding objects.
    \item We construct the GoGetIt and TextCaps-test benchmarks, which contain navigation/manipulation instructions as well as images containing scene text.
    \item In real-world hardware experiments with zero-shot transfer, STARE achieves an overall success rate of 80\%.
\end{itemize}

\section{Related Work}
Vision-language models (VLMs) for robotics and multimodal retrieval are extensively studied~\cite{brohan2023saycan, brohan2022rt} and systematically reviewed~\cite{XIAO2025129963, wang2024crossmodalretrievalsystematicreview}.
Furthermore, integrating scene text is effective for disambiguating visual representations~\cite{Long2018SceneTD, Gupta2022}.

\textbf{Foundation Models for Robotics.}
Recent studies show that remarkable generalization across manipulation and navigation tasks can be achieved by leveraging large-scale foundation models, such as large language models (LLMs) and VLMs~\cite{brohan2023saycan, Driess2023PaLMEAE}. 
Vision-language-action models~\cite{brohan2022rt,team2025gemini} trained on large-scale real robot trajectories enable scalable policy learning across diverse manipulation tasks.
However, these approaches are not well suited for scenarios involving thousands of input images, as such large-scale processing incurs substantial inference latency and often exceeds API token limits.

\textbf{Multimodal retrieval.}
Numerous studies~\cite{Yenamandra2023HomeRobotOM, robocup} have assessed the performance of domestic service robots in mobile manipulation tasks based on natural language instructions.
Several existing methods~\cite{nlmap, Sigurdsson2023RRExBoTRR, dm2rm, relaxformer} perform multimodal retrieval to identify target objects in the environment, and output ranked lists of relevant images based on natural language instructions.
Other methods retrieve target objects and output text-based answers or waypoints~\cite{xie2025embodiedrag}.
Recent benchmarks also evaluate multimodal retrieval in terms of reasoning and knowledge~\cite{lin2026ark}.
Despite rapid advances, few methods leverage scene text (e.g., labels, signage), although it often disambiguates the target.
We address this gap through scene text-aware multimodal retrieval.

\textbf{Scene Text.}
Research on images containing scene text includes recognition and detection~\cite{Liang_2024_CVPR}, retrieval~\cite{Zeng2024FocusDA, vista}, and VQA/synthesis~\cite{Biten2021LaTrLT, Cui_2024_CVPR}.
In particular, scene text retrieval methods often integrate OCR-extracted textual and positional information with visual features to improve alignment with query texts and achieve higher retrieval accuracy.
Mishra et al.~\cite{ocrvqa} study visual question answering by reading text in images.
ViSTA~\cite{vista} embeds OCR-extracted text and positions and fuses them with visual features via a fusion token in a transformer.
In contrast to these approaches, STARE integrates text and position into a narrative representation of object-level associations and introduces an auxiliary similarity function for efficient matching.

\section{Methodology}

\begin{figure*}[t]
    \centering
    \includegraphics[clip,width=\linewidth]{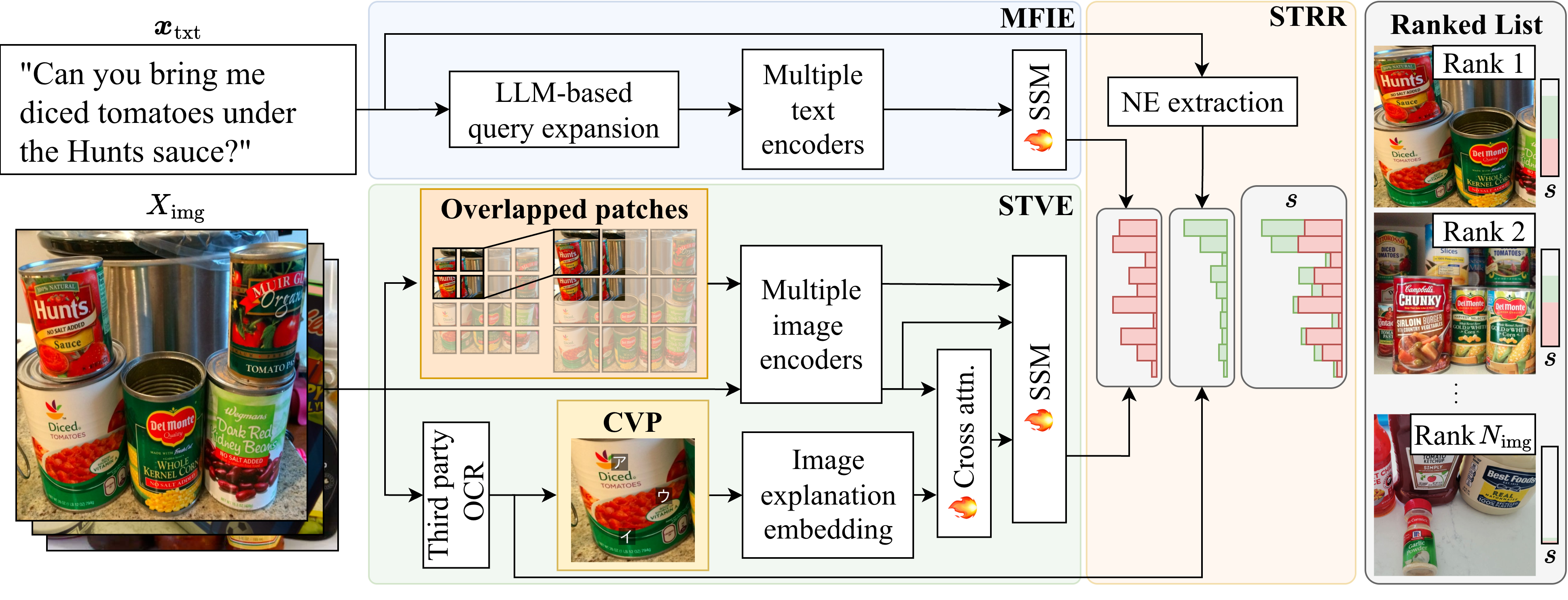}
    \caption{\textbf{Architecture of STARE}.
    STARE includes a narrative representation that integrates Crosslingual Visual Prompt (CVP) and scene text, and reranks based on lexical matches between named entities in the instruction and scene texts.
    SSM represents a Mamba encoder.
    }
    \label{fig:model}
    \vspace{-2mm}
\end{figure*}

\subsection{Problem Statement}
In this paper, we define the Scene Text-Aware Multimodal Retrieval (STMR) task as follows:
a mobile robot collects images, which may include scene texts, from a wide range of indoor or outdoor environments through pre-exploration. 
Given a free-form natural language instruction to {\color{fix}manipulate an object or navigate}, the model retrieves the target images from the {\color{fix}pre-}collected images.
{\color{fix}STARE does not assume the presence of scene text and is applicable whether or not scene text is present, as demonstrated by its performance on the LTRRIE benchmark.}
The model is expected to rank images that correspond to the instruction highly in the output list.
Additional task details are provided in Section~\ref{sec:task_details} of the supplementary material.


The terminology used in this paper is defined as follows.
An ``instruction'' refers to a free-form natural language instruction for object manipulation or navigation that includes referring expressions.
A ``target object'' is the object referred to by an instruction, and a ``target image'' is an image containing that object.
An instruction may have multiple target images.
``Scene text'' refers to text that exists within an image.
We assume that the target images contain either objects that serve as navigation targets (e.g., buildings, signs) or everyday objects that can be manipulated (e.g., kitchen utensils, bottles).

\subsection{Overview of STARE}
We propose STARE, which consists of three modules: the Multi-Form Instruction Encoder (MFIE), Scene Text-aware Visual Encoder (STVE), and Scene Text ReRanker (STRR).
Fig.~\ref{fig:model} shows the architecture of STARE.
STARE separates image-side processing during pre-exploration from instruction-side processing at query time.
During pre-exploration, OCR and CVP are first applied to each candidate image, and STVE generates and stores its image embedding.
At query time, query expansion and named-entity extraction are applied to the instruction, after which MFIE generates the instruction embedding.
The instruction embedding is compared with the precomputed image embeddings, and STRR further reranks the candidates using lexical matches between named entities in the instruction and scene text in each image.
Finally, the images are sorted according to the resulting similarity scores to produce the ranked list.
The input $\bm{x}$ to STARE is defined as follows: 
\begin{align}
\bm{x}=\{\bm{x}_{\text{txt}}, X_{\text{img}}\}, \quad X_{\text{img}}=\{\bm{x}_{\text{img}}^{(i)}\}_{i=1}^{N_{\text{img}}}, 
\end{align}
where $\bm{x}_{\text{img}}^{(i)} \in \mathbb{R}^{3 \times W \times H}$, $\bm{x}_{\text{txt}}$, and $N_\text{img}$ represent an image, an instruction, and the number of candidate images for retrieval, respectively.
Here, $W$ and $H$ denote the common image width and height, respectively.

\subsection{Multi-Form Instruction Encoder}
\label{sec:mfie}
At query time, STARE first performs query-side preprocessing on $\bm{x}_{\text{txt}}$.
Query expansion and named-entity extraction are generated jointly in a single LLM call.
The expanded query is used by MFIE, whereas the extracted named entities are subsequently used by STRR.
MFIE models complex expressions within $\bm{x}_{\text{txt}}$ by incorporating 
text embeddings and LLM-based query expansion.
Query expansion, a technique for expanding a query with semantically related terms, has been less explored in closely related studies.
We do not use $\bm{x}_{\text{txt}}$ directly, because named entities in $\bm{x}_{\text{txt}}$ (e.g., ``Thayers Witch Hazel'') are often underrepresented in the training data of VLMs, which can result in inadequate multimodal embeddings.
Instead, we apply query expansion to generate descriptions capturing the visual characteristics of named entities (e.g., ``A transparent glass bottle with a red lid and ...'').
{\color{fix}Query expansion thus aligns named entities with familiar visual attributes. It uses the LLM's world knowledge to describe the target's appearance, complementing lexical matching when product names are incomplete, paraphrased, or translated.}

Based on $\bm{x}_{\text{txt}}$, MFIE leverages a language encoder and an image-aligned text encoder to extract various features associated with 
$\bm{x}_{\text{txt}}$
at multiple semantic levels.
The unified representation $\bm{h}_{\text{txt}} \in \mathbb{R}^{d_{\text{txt}}}$ is obtained as follows:
\begin{align}
\bm{h}_{\text{txt}} = \text{SSM}\left(
\left[
f(\bm{x}_{\text{txt}});
f(\text{QE}(\bm{x}_{\text{txt}}))
\right]
\right),
\end{align}
where $d_{\text{txt}}$ denotes the output dimension of the MFIE module, and $f(\cdot)$ represents multiple text encoders, such as Stella~\cite{stella} and the CLIP text encoder~\cite{clip}.
$\text{QE}(\bm{x}_{\text{txt}})$ is the LLM-based query expansion of $\bm{x}_{\text{txt}}$.
Stella extracts pure linguistic features, whereas the CLIP text encoder produces image-aligned textual representations.
{\color{fix}We use an SSM to model relationships among multi-granularity features and to improve inference efficiency.}

\subsection{Scene Text-aware Visual Encoder}
\label{sec:stve}
STVE combines image features extracted using image encoders, fine-grained image features using overlapped patchification, and narrative representations that consider scene text.
The overlapped patchified features allow the module to capture finer details of the image.
Overlapped patchification is used because standard patchification cannot accurately capture visual features of objects or scene texts located at patch boundaries.
Specifically, $\bm{x}_{\text{img}}^{(i)}$ is divided into overlapping cells, which are encoded by multiple image encoders (e.g., CLIP, DINOv2~\cite{dinov2}) to produce multimodal features $\bm{h}_{\text{fine}}^{(i)}$.
Using multiple image encoders allows the model to capture information across both semantic and spatial levels, as shown in~\cite{shi2025eagle}.

Next, we introduce the narrative representations, which are generated using Crosslingual Visual Prompt (CVP).
Many everyday objects contain scene texts, making it essential to extract image features that effectively incorporate scene texts.
However, existing methods \cite{nlmap, beit3, dm2rm, relaxformer} do not sufficiently integrate scene texts with image features, limiting their effectiveness in multimodal retrieval tasks.
To fully leverage scene texts, simply extracting scene texts and their positions using OCR is insufficient.
In environments where multiple scene texts are present, OCR results are not enough to accurately associate each text with the correct object.
Occlusions or unfavorable orientations can also result in inaccurate OCR-detected scene texts.
To address these challenges, we use an MLLM to extract linguistic features from images that capture complex referring expressions, spatial relationships, and relationships between objects and scene text.

However, relying solely on MLLMs can cause hallucinations, where incorrect or nonexistent text is generated.
We mitigate this issue by providing the MLLM with OCR-detected scene texts and visually marking their locations within the image.
We do not use conventional visual prompting because inserted prompts may be misinterpreted as scene text by downstream multimodal models.
Unlike typical visual prompting, {\color{fix}CVP employs non-alphanumeric visual marks to distinguish the marks from scene texts present in the image.}
These marks are chosen for their suitability for enumerated list-style descriptions.

This representation is generated by first performing third-party OCR
on $\bm{x}_{\text{img}}^{(i)}$ to extract 
scene texts
along with their corresponding locations within the image.
Using the OCR results, each scene text in the image is enclosed within a bounding box, and a crosslingual mark is overlaid at the top of the bounding box to create CVP.
The language used for the CVP marks is described in Section~\ref{quantitative_results}.
The CVP and scene texts
are then passed to a multimodal large language model (MLLM) to generate a description of the image that considers the scene text.
{\color{fix}Here, the MLLM is used only to generate image captions from the provided CVP and is not instructed to inject external knowledge.}
The detailed motivation and the prompt used by the ``Image explanation embedding'' in Fig.~\ref{fig:model} are provided in Section~\ref{sec:method_details_stve} of the supplementary material.

The description is then passed to a language encoder to create the narrative representation $\bm{l}_{\text{N}}^{(i)}$.
We integrate $\bm{l}_{\text{N}}^{(i)}$ with pure visual representations to obtain multimodal features, which can be used in parallel with other multimodal representations.
$\bm{l}_{\text{N}}^{(i)}$ encodes information ranging from low-level attributes (e.g., color, texture) to high-level concepts (e.g., product names).
To match this range, we use visual features from multiple intermediate layers of DINOv2.


The final output of the STVE module $\bm{h}_{\text{img}}^{(i)} \in \mathbb{R}^{d_{\text{img}}}$ can be calculated as follows: 
\begin{align}
\bm{h}_{\text{img}}^{(i)} = \text{SSM}([
\bm{h}_{\text{coarse}}^{(i)}; 
\bm{h}_{\text{fine}}^{(i)};
\text{CrossAttn}(\bm{l}_{\text{N}}^{(i)}, \bm{v}_{\text{coarse}}^{(i)})
]
),
\end{align}
where $\bm{h}_{\text{coarse}}^{(i)}$, $\bm{v}_{\text{coarse}}^{(i)}$, and $d_{\text{img}}$ denote the coarse multimodal features extracted from $\bm{x}_{\text{img}}^{(i)}$ using multiple image encoders, the coarse visual features derived from the intermediate layers of DINOv2 for $\bm{x}_{\text{img}}^{(i)}$, and the output dimension of the STVE module, respectively.

\subsection{Scene Text ReRanker}
\label{sec:strr}
We introduce STRR to model the relationships between named entities in the instruction and scene text associated with candidate objects.
This is particularly important in the STMR task, where product names, store names, and other named entities can provide strong cues for identifying the corresponding objects.
Existing methods typically score the $\bm{h}_{\text{txt}}$ and $\bm{h}_{\text{img}}^{(i)}$ pairs using cosine similarity~\cite{clip}, which does not directly capture lexical matches between named entities and scene text.
Moreover, retrieval settings require fast similarity calculations using pre-extracted features.

Thus, we introduce a similarity score $s(\cdot,\cdot)$ that combines cosine similarity with an additional term for matching named entities against scene text:
\belowdisplayskip = 1mm
\begin{equation}
\label{eq:strr_similarity}
\begin{aligned}
s(\bm{x}_{\text{txt}}, \bm{x}_\text{img}^{(i)}) &=
\frac{\bm{h}_{\text{txt}} \cdot \bm{h}_{\text{img}}^{(i)}}{\|\bm{h}_{\text{txt}}\| \|\bm{h}_{\text{img}}^{(i)}\|}
+ \lambda \rho_i, \\
\rho_i &=
\begin{cases}
\dfrac{|T_{\text{NE}} \cap T_{\text{ST}}^{(i)}|}{\max_j |T_{\text{NE}} \cap T_{\text{ST}}^{(j)}|}, & \text{if } \max_j |T_{\text{NE}} \cap T_{\text{ST}}^{(j)}| > 0, \\
0, & \text{otherwise}.
\end{cases}
\end{aligned}
\end{equation}
where $\lambda$ is a learnable weight that controls the influence of OCR-based reranking.
When an image has no matching named-entity and scene-text tokens, the reranking term is zero and the score reduces to cosine similarity.
Here, $T_{\text{NE}}$ and $T_{\text{ST}}^{(i)}$ denote the token sets of named entities and scene texts, respectively, generated using a tokenizer (e.g., tiktoken).
The named entities are extracted from $\bm{x}_\text{txt}$ using an LLM (denoted as NE extractor in Fig.~\ref{fig:model}).
To address the significant amount of noise in scene texts, we enhance the robustness of our method by applying preprocessing techniques such as normalization (e.g., case normalization, symbol removal) in addition to simple string matching.
The incentive term in the similarity score evaluates lexical matches between named entities and scene texts,
while retaining fast matching capabilities.
The usage of $T_{\text{NE}}$ enables reranking that prioritizes semantically relevant matches.
The final ranked image list $\hat{Y}$ is obtained by sorting $X_{\text{img}}$ based on $s(\bm{x}_{\text{txt}}, \bm{x}_\text{img}^{(i)})$.
\section{Experiments}
\label{sec:experimental_results}

\subsection{Experimental Setup}
\label{sec:experimental_setup}

\textbf{Benchmarks.}
We constructed the GoGetIt and TextCaps-test benchmarks for the STMR task.
The GoGetIt benchmark was constructed because existing multimodal retrieval datasets~\cite{textcaps, stacmr} are not always suitable for robotic manipulation or navigation tasks.
Although the RefText dataset~\cite{stan} uses scene texts to identify the target object, some instructions are overly simple, with several consisting of only a single word.
To address this, we annotated samples with longer and more practical instructions that describe robotic actions suitable for execution in real-robot experiments.

However, evaluating multimodal foundation models on RefText raises potential data-leakage concerns because some RefText images originate from public datasets such as COCO~\cite{Lin2014MicrosoftCC}, which may also be used during pretraining.
RefText's random train/validation/test split~\cite{stan} adds to these concerns.
{\color{fix}This concern is substantial: since 61.9\% of RefText samples are from public datasets, approximately 50\% may overlap with the pretraining corpora of baselines like Long-CLIP and BLIP-2.}
Thus, we required a different benchmark to enable a fair comparison with {\color{fix}foundation-model} baselines.
The original TextCaps captions are not robot instructions, so we created TextCaps-test from TextCaps~\cite{textcaps} with newly annotated instructions.

%

We also evaluated the performance of our method in settings without scene text using the LTRRIE dataset~\cite{multirankit}, a multimodal retrieval dataset featuring indoor environments.
The instructions and images in the LTRRIE dataset were collected from the REVERIE dataset~\cite{qi2020reverie} and the Matterport3D simulator, respectively.

\textbf{Statistics.}
The GoGetIt benchmark consists of 27,280 samples and 3,657 images containing scene text, collected from diverse indoor and outdoor environments.
The GoGetIt-RefText subset contains 23,323 samples with instructions sourced from the RefText~\cite{stan} dataset, while the GoGetIt-Instruction subset contains 3,957 referring-expression instructions newly collected from 129 annotators.
The TextCaps-test benchmark consists of 400 instructions provided by 10 annotators and 105 images containing scene text.
The LTRRIE~\cite{multirankit} dataset consists of 5,108 English instructions and 3,569 images, collected from 67 indoor environments.
{\color{fix}For LTRRIE, we use the original train/validation/test split and image counts.}

In the GoGetIt benchmark, the RefText and Instruction subsets were partitioned following the same split as the RefText dataset.
The training set contains 17,352 samples from the RefText subset and 3,126 from the Instruction subset.
The validation set contains 1,433 RefText samples and 266 Instruction samples.
The test set includes 4,538 RefText samples, 565 Instruction samples, and 400 TextCaps-test samples.
Including LTRRIE, the training set consists of 31,071 samples, the validation set contains 2,096 samples, and the test set includes 6,004 samples.
There is no overlap of samples across the training, validation, and test sets.
We used the training set to train the model, the validation set to tune hyperparameters, and the test set to evaluate model performance.

\textbf{Annotation.}
For the GoGetIt-Instruction subset and TextCaps-test benchmark, annotators were shown each image with the target object enclosed in a bounding box, together with scene text extracted from the image.
They were asked to provide full-sentence instructions describing object manipulation or navigation actions, optionally including scene text.
{\color{fix}Multiple experts manually reviewed all test samples against predefined criteria. Each instruction had to be a full-sentence manipulation or navigation command consistent with the target object and its scene text.}

\Needspace{17\baselineskip}
\begin{wraptable}{r}{0.42\textwidth}
    \vspace{-3mm}
    \centering
    \footnotesize
    \setlength{\tabcolsep}{3pt}
    \caption{Experimental settings for our method.}
    \label{tab:parameters}
    \begin{tabular}{lc}
        \hline
        Optimizer & AdamW \\
        $\beta_1$, $\beta_2$ & 0.9, 0.98 \\
        Learning rate & $1\times10^{-4}$ \\
        Batch size & 128 \\
        Number of epochs & 20 \\
        $L_\text{st}$, $H_\text{st}$, $A_\text{st}$ & 5, 2,048, 1 \\
        $L_\text{np}$, $H_\text{np}$, $A_\text{np}$ & 4, 2,304, 8 \\
        $d_\text{txt}$, $d_\text{img}$ & 1024, 1024 \\
        \hline
    \end{tabular}
\end{wraptable}

\textbf{Training Setup.}
\label{sec:stare_training_setup}
Table~\ref{tab:parameters} summarizes the optimizer, learning rate, batch size, number of epochs, and encoder settings used to train STARE.
Here, $L_\text{st}$, $H_\text{st}$, and $A_\text{st}$ denote the layer count, hidden dimension, and attention-head count of the Scene Text-aware Visual Encoder module, while $L_\text{np}$, $H_\text{np}$, and $A_\text{np}$ denote the corresponding quantities for the Multi-Form Instruction Encoder module.
Details of the loss function and evaluation metric are provided in Sections~\ref{sec:loss_function} and~\ref{sec:evaluation_metric} of the supplementary material, respectively.

\subsection{Implementation Details}
\label{sec:implementation_details}

We used GPT-4o~\cite{gpt_4o} for MFIE and STVE, and Google Cloud Vision API for OCR.
During pre-exploration, OCR, backbone feature extraction, and STVE are executed once for the candidate images, and the resulting image embeddings are stored.
At query time, only instruction-side processing and STRR are executed.
STARE had approximately 40 million trainable parameters and 5 billion multiply--add operations.
We trained our model on a GeForce RTX 4090 with 24 GB of VRAM and an Intel Core i9-13900KF with 64 GB of CPU memory.
Training took approximately 1.2 hours.
The similarity calculation between a single instruction and 100 precomputed image embeddings took approximately 39.8 ms.
We measured recall@10 on the validation set for each epoch.
The evaluation on the test set was performed using the model that had the highest recall@10 score on the validation set.
We used the Double Relaxed Contrastive loss~\cite{relaxformer} with the STRR score $s$, including the lexical matching term.

\textbf{Baselines.}
{\color{fix}
Our zero-shot multimodal retrieval baselines were CLIP (ViT-L/14), Long-CLIP (ViT-L/14), BLIP-2 (ViT-g), BEiT-3 (large), and Meta CLIP 2~\cite{clip, longclip, blip2, beit3, chuang2025metaclip2}.
We also included ReMEmbR, NLMap, and RelaX-Former due to their demonstrated effectiveness in image retrieval and retrieval-based fetch-and-carry tasks~\cite{anwar2025remembr, nlmap, relaxformer}.
In addition, we fine-tuned the text and image encoders of CLIP and Long-CLIP as extra baselines.
}
Additional recall@20 results are reported in Section~\ref{sec:additional_quantitative_results} of the supplementary material.

\begin{table*}[t]
\centering
\caption{Quantitative comparison on GoGetIt-RefText, GoGetIt-Instruction, TextCaps-test, and LTRRIE.
Bold, underline, `*', and `FT' denote the best score, second-best score, reproduced results, and fine-tuning, respectively.
BLIP-2, Long-CLIP, and RelaX-Former scores on GoGetIt are omitted because of training--test data leakage (Section~\ref{sec:experimental_setup}).
}
\resizebox{\textwidth}{!}{
\begin{tabular}{lccccccccccccc}
\toprule
\hspace{-2pt}[\%]
& \multirow{2}{*}{FT}
& \multicolumn{2}{c}{GoGetIt (RefText)}  
& \multicolumn{2}{c}{GoGetIt (Instruction)}  
& \multicolumn{2}{c}{TextCaps-test}  
& \multicolumn{2}{c}{LTRRIE} 
\\ 
\cmidrule(lr){3-4}
\cmidrule(lr){5-6}
\cmidrule(lr){7-8}
\cmidrule(lr){9-10}
Method
&
& \small R@5 $\uparrow$
& \small R@10 $\uparrow$
& \small R@5 $\uparrow$
& \small R@10 $\uparrow$
& \small R@5 $\uparrow$
& \small R@10 $\uparrow$
& \small R@5 $\uparrow$
& \small R@10 $\uparrow$

\\ 
\midrule 
CLIP~\cite{clip}
& \textcolor{cyan}{\faSnowflake}
& 63.2
& 72.8
& 74.3
& 83.3 
& 79.8
& 86.3
& 56.1
& 71.0
\\

CLIP~\cite{clip}
& \textcolor{red}{\faFire}
& 63.5\footnotesize~$\pm$~0.3
& 74.1\footnotesize~$\pm$~0.3
& 73.9\footnotesize~$\pm$~0.7
& 83.9\footnotesize~$\pm$~0.3
& 82.0\footnotesize~$\pm$~0.6
& 90.1\footnotesize~$\pm$~0.4
& 56.8\footnotesize~$\pm$~0.7
& 72.3\footnotesize~$\pm$~0.4
\\

NLMap* \cite{nlmap}
& \textcolor{cyan}{\faSnowflake}
& 50.9
& 60.1 
& 61.0 
& 73.5 
& 70.0
& 78.8 
& 50.9
& 66.4
\\

BEiT-3 \cite{beit3}
& \textcolor{cyan}{\faSnowflake}
& 54.4
& 65.3
& 63.7 
& 79.5 
& 76.5
& 84.8
& 59.9
& 76.6 
\\

BLIP-2 \cite{blip2}
& \textcolor{cyan}{\faSnowflake}
& - 
& - 
& - 
& - 
& 76.5
& 86.3
& 66.6
& 81.9
\\

Long-CLIP~\cite{longclip}
& \textcolor{cyan}{\faSnowflake}
& -
& -
& -
& -
& 86.0
& 90.3
& 61.6
& 79.3
\\

Long-CLIP~\cite{longclip}
& \textcolor{red}{\faFire}
& - 
& - 
& - 
& - 
& 87.2\footnotesize~$\pm$~0.1
& \uline{91.8}\footnotesize~$\pm$~0.3
& \uline{69.9}\footnotesize~$\pm$~0.3
& \uline{84.3}\footnotesize~$\pm$~0.1
\\

RelaX-Former~\cite{relaxformer}
& \textcolor{red}{\faFire}
& - 
& - 
& - 
& - 
& 62.3\footnotesize~$\pm$~1.2
& 73.7\footnotesize~$\pm$~0.7
& 66.6\footnotesize~$\pm$~0.9
& 81.7\footnotesize~$\pm$~0.7
\\

ReMEmbR* \cite{anwar2025remembr}
& \textcolor{cyan}{\faSnowflake}
& 40.8 
& 48.7 
& 50.3 
& 58.7 
& 71.0 
& 78.3 
& 27.1
& 39.0
\\

Meta CLIP 2 \cite{chuang2025metaclip2}
& \textcolor{cyan}{\faSnowflake}
& 70.4
& 78.1
& \uline{77.5}
& \uline{84.6}
& \uline{87.5}
& 91.5
& 61.5
& 78.5
\\

ViSTA* \cite{vista}
& \textcolor{red}{\faFire}
& \uline{72.4}\footnotesize~$\pm$~0.5
& \uline{81.2}\footnotesize~$\pm$~0.4
& 72.7\footnotesize~$\pm$~0.6
& 82.5\footnotesize~$\pm$~0.5
& 78.2\footnotesize~$\pm$~0.7
& 86.5\footnotesize~$\pm$~0.4
& 62.6\footnotesize~$\pm$~0.8
& 80.3\footnotesize~$\pm$~0.6
\\

STARE (ours)
& \textcolor{red}{\faFire}
& \textbf{90.2}\footnotesize~$\pm$~0.4
& \textbf{94.4}\footnotesize~$\pm$~0.4
& \textbf{89.8}\footnotesize~$\pm$~0.6
& \textbf{92.8}\footnotesize~$\pm$~0.6
& \textbf{92.4}\footnotesize~$\pm$~1.3
& \textbf{95.4}\footnotesize~$\pm$~0.6
& \textbf{72.5}\footnotesize~$\pm$~0.8
& \textbf{86.6}\footnotesize~$\pm$~0.9
\\ 

\bottomrule
\end{tabular}
}

\label{tab:results}
\vspace{-2mm}
\end{table*}

\subsection{Quantitative Results}
\label{quantitative_results}
%
Table~\ref{tab:results} presents a quantitative comparison of STARE and the baseline methods.
We used recall@$K$ ($K$=5,10), a standard metric for multimodal retrieval, and treated recall@10 as the primary measure~\cite{Cao+ijcal22}.
Scores for the frozen baselines are reported from a single trial because their scores were consistent across repeated runs.
For the other methods, we report the mean and standard deviation obtained from five trials.
The highest score for each metric is in bold.
The scores for Long-CLIP, BLIP-2, and RelaX-Former are not reported on the GoGetIt benchmark due to data leakage between their training data and the test set of the GoGetIt benchmark.
We used Katakana characters as the non-alphanumeric characters for CVP, which are visually distinguishable from the scene texts (see Table~\ref{tab:ablation3}).

\begin{figure*}[t]
    \centering
    \includegraphics[clip,width=0.92\linewidth]{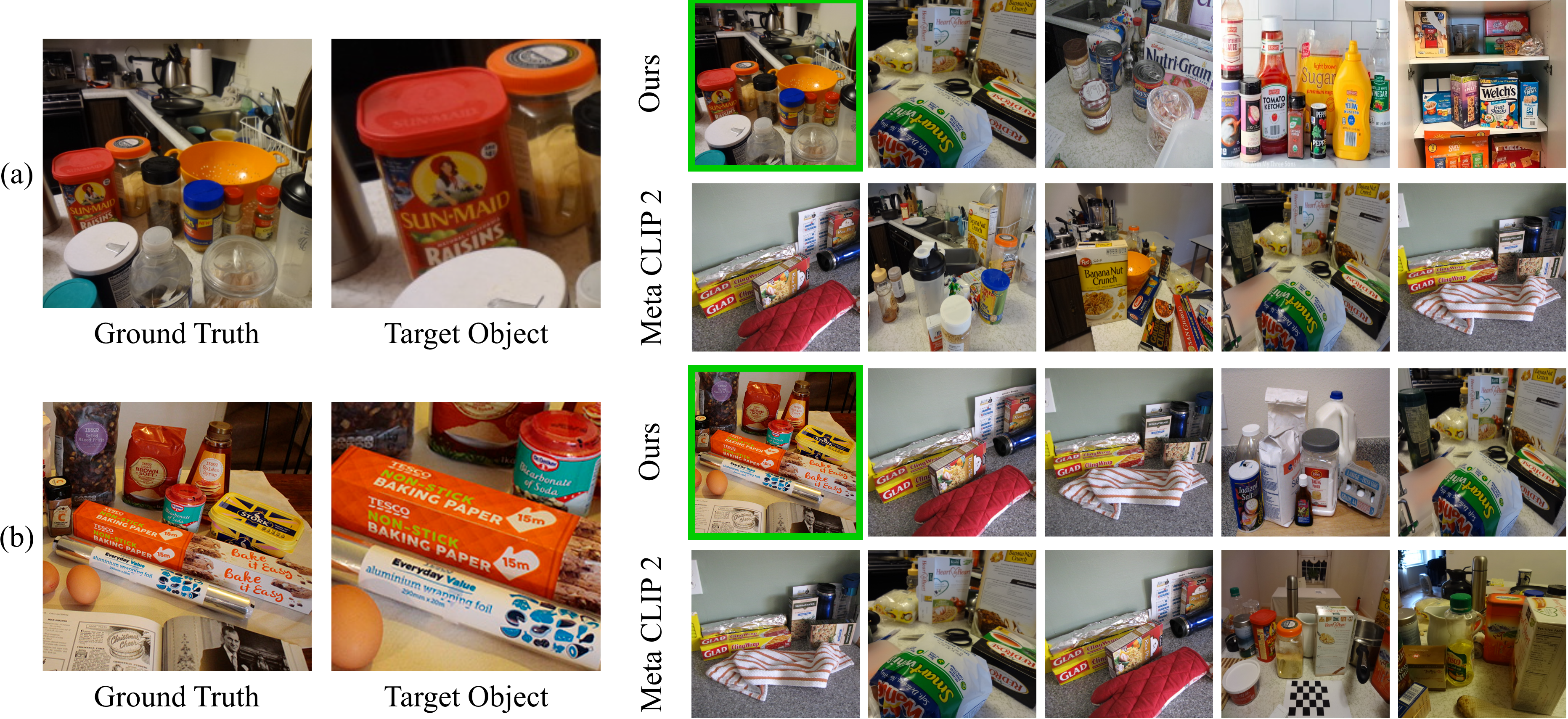}
    \caption{\textbf{Qualitative results} for STARE and Meta CLIP 2~\cite{chuang2025metaclip2} on (a) GoGetIt-Instruction and (b) GoGetIt-RefText. The instructions are (a) ``Pass me the red container of Sun-Maid raisins on the kitchen counter.'' and (b) ``A long box of non-stick baking paper.'' Green borders mark target images in the retrieval lists.}
    \label{fig:success}
    \vspace{-2mm}
    
\end{figure*}

Table~\ref{tab:results} shows that STARE achieved the highest recall@10 scores of 94.4\%, 92.8\%, 95.4\%, and 86.6\% on the RefText subset of the GoGetIt benchmark, the Instruction subset of the GoGetIt benchmark, the TextCaps-test benchmark, and the LTRRIE dataset, respectively. 
Our method outperformed the best baseline scores of 81.2\%, 84.6\%, 91.8\%, and 84.3\% by 13.2, 8.2, 3.6, and 2.3 points, respectively.
Moreover, STARE outperformed all baseline methods across all other evaluation metrics.
These differences in recall@10 were statistically significant ($p < 0.01$).
%

\subsection{Qualitative Results}
Fig.~\ref{fig:success} shows qualitative results of STARE and Meta CLIP 2, where (a) and (b) present samples from the Instruction subset and RefText subset of the GoGetIt benchmark, respectively.
For each sample, the target image and the top-5 images retrieved by STARE and the baseline method are shown.
The target image is highlighted with a green border.

Fig.~\ref{fig:success} (a) presents an example for which $\bm{x}_{\text{txt}}$ was ``Pass me the red container of Sun-Maid raisins on the kitchen counter,'' and the target object was the red container with the scene text ``Sun-Maid raisins''.
While the baseline method ranked images unrelated to Sun-Maid raisins in the top-5, STARE correctly placed the target image as rank 1.

In Fig.~\ref{fig:success} (b), $\bm{x}_{\text{txt}}$ was ``a long box of non-stick baking paper.''
The baseline method ranked images that did not include baking paper in the top-5.
In contrast, the proposed method ranked the target image, which includes a box labeled ``non-stick baking paper,'' as the top result.

These examples exhibit STARE’s ability to retrieve images based on instructions containing scene texts, spatial relations between objects, and referring expressions.
A failure case and error analysis are provided in Sections~\ref{sec:additional_qualitative_results} and~\ref{sec:error_analysis} of the supplementary material, respectively.
\subsection{Ablation Studies}
\begin{table*}[t]
\centering
\caption{
\textbf{Ablation results}. Mean and standard deviation over five trials are reported; bold indicates the highest value. Models (a)--(i), (k), and (l) are shown. K, AN, E, G, and BB denote Katakana, Arabic numerals, English, Greek, and bounding box, respectively.
}
\label{tab:ablation3}
\resizebox{\textwidth}{!}
{
\small
\renewcommand{\arraystretch}{1.15}
\begin{tabular}{lccccccccccc}
\toprule
\hspace{-2pt}{[\%]}
& \multirow{2}{*}{STRR} 
& \multirow{2}{*}{CVP} 
& \multirow{2}{*}{Patchification} 
& \multicolumn{2}{c}{GoGetIt(RefText)}  
& \multicolumn{2}{c}{GoGetIt(Instruction)}  
& \multicolumn{2}{c}{TextCaps-test}  
& \multicolumn{2}{c}{LTRRIE} 
\\ 
\cmidrule(lr){5-6}
\cmidrule(lr){7-8}
\cmidrule(lr){9-10}
\cmidrule(lr){11-12}
Model
& 
&
&
& R@5 $\uparrow$  
& R@10 $\uparrow$ 
& R@5 $\uparrow$  
& R@10 $\uparrow$
& R@5 $\uparrow$  
& R@10 $\uparrow$   
& R@5 $\uparrow$  
& R@10 $\uparrow$

\\
\hline
(a)  
& $\boldsymbol{T}_{\text{NE}}$
& K
& overlap
& \textbf{90.2}\scriptsize~$\pm$~0.4
& \textbf{94.4}\scriptsize~$\pm$~0.4
& \textbf{89.8}\scriptsize~$\pm$~0.6
& \textbf{92.8}\scriptsize~$\pm$~0.6
& 92.4\scriptsize~$\pm$~1.3
& \textbf{95.4}\scriptsize~$\pm$~0.6
& \textbf{72.5}\scriptsize~$\pm$~0.8
& \textbf{86.6}\scriptsize~$\pm$~0.9

\\
\hline

(b) 
& - 
& K
& overlap
& 76.4\scriptsize~$\pm$~0.8
& 85.0\scriptsize~$\pm$~0.7
& 78.2\scriptsize~$\pm$~1.2
& 85.9\scriptsize~$\pm$~0.5
& 83.6\scriptsize~$\pm$~1.1
& 89.5\scriptsize~$\pm$~1.9
& 71.1\scriptsize~$\pm$~1.0
& 85.1\scriptsize~$\pm$~0.6
\\

(c)  
& $\boldsymbol{T}_{\text{txt}}$
& K
& overlap
& 81.4\scriptsize~$\pm$~0.8
& 89.2\scriptsize~$\pm$~0.6
& 79.7\scriptsize~$\pm$~1.1
& 87.7\scriptsize~$\pm$~1.1
& 82.1\scriptsize~$\pm$~0.6
& 89.4\scriptsize~$\pm$~1.0
& 70.6\scriptsize~$\pm$~1.3
& 85.1\scriptsize~$\pm$~0.8
\\

\hline

(d)  
& $\boldsymbol{T}_{\text{NE}}$
& AN
& overlap
& \textbf{90.2}\scriptsize~$\pm$~0.6
& 94.3\scriptsize~$\pm$~0.2
& 88.4\scriptsize~$\pm$~0.5
& 92.3\scriptsize~$\pm$~0.8
& 92.3\scriptsize~$\pm$~1.0
& 95.3\scriptsize~$\pm$~0.4
& 71.3\scriptsize~$\pm$~1.0
& 85.9\scriptsize~$\pm$~0.3
\\

(e)
& $\boldsymbol{T}_{\text{NE}}$
& E
& overlap
& 90.1\scriptsize~$\pm$~0.9
& 94.3\scriptsize~$\pm$~0.4
& 89.3\scriptsize~$\pm$~0.5
& 92.2\scriptsize~$\pm$~0.6
& 91.3\scriptsize~$\pm$~0.6
& 95.3\scriptsize~$\pm$~0.6
& 71.6\scriptsize~$\pm$~0.9
& 85.7\scriptsize~$\pm$~1.3
\\

(f)
& $\boldsymbol{T}_{\text{NE}}$
& G
& overlap
& 90.1\scriptsize~$\pm$~0.8
& 94.3\scriptsize~$\pm$~0.3
& 88.9\scriptsize~$\pm$~0.4
& 92.1\scriptsize~$\pm$~0.5
& 91.9\scriptsize~$\pm$~0.9
& 95.3\scriptsize~$\pm$~0.5
& 71.2\scriptsize~$\pm$~1.0
& 86.0\scriptsize~$\pm$~0.5
\\

(g)  
& $\boldsymbol{T}_{\text{NE}}$
& BB
& overlap
& 90.0\scriptsize~$\pm$~0.2
& 94.2\scriptsize~$\pm$~0.2
& 89.1\scriptsize~$\pm$~0.8
& 92.6\scriptsize~$\pm$~0.5
& 91.9\scriptsize~$\pm$~1.1
& 95.2\scriptsize~$\pm$~0.1
& 70.9\scriptsize~$\pm$~2.0
& 85.5\scriptsize~$\pm$~1.4
\\

\hline

(h) 
& $\boldsymbol{T}_{\text{NE}}$
& -
& overlap
& 87.0\scriptsize~$\pm$~0.4
& 92.1\scriptsize~$\pm$~0.5
& 87.3\scriptsize~$\pm$~0.6
& 91.9\scriptsize~$\pm$~0.5
& 89.5\scriptsize~$\pm$~1.0
& 93.7\scriptsize~$\pm$~0.5
& 67.2\scriptsize~$\pm$~1.5
& 82.3\scriptsize~$\pm$~0.7
\\

(i) 
& -
& -
& overlap
& 69.3\scriptsize~$\pm$~1.1
& 79.7\scriptsize~$\pm$~0.5
& 73.6\scriptsize~$\pm$~0.8
& 82.3\scriptsize~$\pm$~1.2
& 77.6\scriptsize~$\pm$~0.9
& 84.8\scriptsize~$\pm$~1.3
& 67.4\scriptsize~$\pm$~0.4
& 81.9\scriptsize~$\pm$~0.6
\\
\hline

(k)
& $\boldsymbol{T}_{\text{NE}}$
& K
& -
& 89.9\scriptsize~$\pm$~0.7
& 94.2\scriptsize~$\pm$~0.6
& 89.5\scriptsize~$\pm$~0.7
& 92.2\scriptsize~$\pm$~0.5
& \textbf{92.5}\scriptsize~$\pm$~0.5
& 95.3\scriptsize~$\pm$~0.1
& 71.3\scriptsize~$\pm$~1.0
& 84.8\scriptsize~$\pm$~0.6
\\

(l)
& $\boldsymbol{T}_{\text{NE}}$
& K
& standard
& 88.4\scriptsize~$\pm$~1.2
& 93.4\scriptsize~$\pm$~0.7
& 88.8\scriptsize~$\pm$~0.6
& 92.4\scriptsize~$\pm$~0.4
& 90.7\scriptsize~$\pm$~0.5
& 94.9\scriptsize~$\pm$~0.6
& 68.3\scriptsize~$\pm$~0.7
& 83.7\scriptsize~$\pm$~0.6
\\

\bottomrule
\end{tabular}
}
\vspace{-2mm}
\end{table*}

Table~\ref{tab:ablation3} presents ablation results for STRR, CVP, OCR-derived features, and patchification.
The average and standard deviation from five trials are reported.
Bold indicates the highest value.

\textbf{STRR Ablation.}
We evaluated the effectiveness of the STRR module by comparing Model (a), our proposed model, with Model (b), which omits the STRR module by replacing it with cosine similarity.
Within models incorporating the STRR module, we further compared Model (a) with Model (c), which uses the entire instruction instead of named entities for reranking.
From the results in Table~\ref{tab:ablation3}, the recall@10 scores on the Instruction subset of the GoGetIt benchmark were 85.9\% for Model (b) and 87.7\% for Model (c).
These scores were 6.9 and 5.1 points lower than the recall@10 score achieved by Model (a), respectively.
The results of this experiment indicate that named entity-based reranking effectively improves retrieval performance by enhancing the consistency between the instructions and the scene texts.

\textbf{CVP Ablation.}
To investigate the effects of the character type on the retrieval performance, we replaced the language used for CVP with Arabic numerals, English, and Greek for Models (d), (e), and (f), respectively.
Model (g) ablates scene texts, using images with only bounding boxes.
From the results in Table~\ref{tab:ablation3}, the recall@10 scores on the Instruction subset of the GoGetIt benchmark were 92.3\%, 92.2\%, 92.1\%, and 92.6\% for Models (d), (e), (f), and (g), respectively, which were 0.5, 0.6, 0.7, and 0.2 points lower than the recall@10 score achieved by Model (a).
These results suggest that using Katakana for CVP effectively aligns other visual features with scene texts while reducing interference with existing scene texts, resulting in improved retrieval performance.

\textbf{OCR Ablation.}
To
assess the impact of
OCR-derived features, we removed OCR-derived features in Models (h) and (i).
Model (h) removes CVP, while (i) removes all OCR-derived features, including third-party OCR, CVP, and STRR.
Results in Table~\ref{tab:ablation3} show that the recall@10 scores on the Instruction subset of the GoGetIt benchmark for Models (h) and (i) were 91.9\% and 82.3\%, respectively, which were 0.9 and 10.5 points lower than that of Model (a).
These results indicate that OCR-derived features improve retrieval performance by explicitly handling scene text.
 

\textbf{Patchification Ablation.}
To determine the contribution of patchification, we removed this functionality in Model (k).
Within models using patchification, the effects of overlapping patches were examined by replacing overlapped patchification with standard patchification in Model (l).
Table~\ref{tab:ablation3} indicates that the recall@10 scores on the Instruction subset of the GoGetIt benchmark were 92.2\% for Model (k) and 92.4\% for Model (l).
These values were 0.6 and 0.4 points lower, respectively, than the score achieved by Model (a).
These results demonstrate that overlapped patchification improves the model’s ability to capture detailed visual features, particularly those located at patch boundaries.
The MFIE ablation is provided in Section~\ref{sec:additional_ablation_study} of the supplementary material.

\subsection{Hardware Experiments}
\textbf{Implementation.}
We validated STARE in real-world zero-shot experiments using a domestic service robot (DSR).
We used everyday objects from the YCB object set~\cite{calli2015ycb}, together with household items bearing printed labels.
The latter allowed us to evaluate performance on objects with scene text.
The objects and the environment used in these experiments were unseen during the training phase.
More detailed setup information is provided in Section~\ref{hardware_setup} of the supplementary material.

\Needspace{20\baselineskip}
\begin{wraptable}{r}{0.50\textwidth}
\vspace{-3mm}

\centering
\captionsetup{skip=1pt}
\caption{Quantitative hardware results. FT: fine-tuned; scores are percentages.}
\label{tab:hardware_results}
{
\footnotesize
\setlength{\tabcolsep}{3pt}
\renewcommand{\arraystretch}{1.0}
\begin{tabular}{@{}lccc@{}}
\toprule
Method & FT & R@5 $\uparrow$ & SR $\uparrow$ \\
\midrule
CLIP~\cite{clip}
& \textcolor{cyan}{\faSnowflake}
& 71           
& 64\\
CLIP~\cite{clip}
& \textcolor{red}{\faFire}
& 74           
& 65\\
NLMap*~\cite{nlmap}  
& \textcolor{cyan}{\faSnowflake}
& 71           
& 65\\
BEiT-3~\cite{beit3}
& \textcolor{cyan}{\faSnowflake}
& 66           
& 62\\
BLIP-2~\cite{blip2}                     
& \textcolor{cyan}{\faSnowflake}
& 72           
& 66\\
Long-CLIP~\cite{longclip}        
& \textcolor{cyan}{\faSnowflake}
& 70           
& 66\\
Long-CLIP~\cite{longclip}
& \textcolor{red}{\faFire}
& 76
& 70\\
RelaX-Former~\cite{relaxformer}             
& \textcolor{red}{\faFire}
& 60           
& 54\\
ReMEmbR*~\cite{anwar2025remembr}                      
& \textcolor{cyan}{\faSnowflake}
& 41          
& 38\\
Meta CLIP 2~\cite{chuang2025metaclip2}
& \textcolor{cyan}{\faSnowflake}
& 80
& 72\\
ViSTA*~\cite{vista}
& \textcolor{red}{\faFire}
& 81
& 73\\
STARE (ours)
& \textcolor{red}{\faFire}
& \textbf{88} 
& \textbf{80}\\
\bottomrule
\end{tabular}
}
\end{wraptable}

In total, 100 episodes were conducted across 12 different environmental setups, with 5-10 episodes per setup.
During pre-exploration, the DSR navigated along 15 waypoints and captured RGB images using the Point Grey Chameleon camera mounted on its head.
The waypoints can be designed using existing viewpoint planning methods\cite{Nilsson2008,Blaer2006}.
The DSR was provided with free-form English instructions, which typically referenced scene texts and involved tasks such as fetching or navigation.

We divided the task into two subtasks: multimodal retrieval and object fetching.
In the multimodal retrieval phase, the baseline methods and STARE were evaluated using the images pre-collected by the robot, and recall@5 was calculated based on the retrieval results.
Object fetching was attempted only when a target image appeared in the top-5 retrieval results.
Grasping points were computed from the depth image and the segmentation mask of the target object.
An episode was considered a success only if the target image was ranked within the top-5 and the robot successfully grasped the corresponding object.
%

\textbf{Quantitative Results.}
STARE was evaluated using recall@5 and success rate (SR),
standard evaluation metrics for multimodal retrieval and robotic manipulation tasks, respectively.
SR is defined as
$\text{SR}=N_\text{s}/N_\text{e}$,
where $N_\text{s}$ and $N_\text{e}$ denote the numbers of successful and evaluated episodes, respectively.

Table~\ref{tab:hardware_results} presents quantitative results of the hardware experiments.
STARE achieved an overall recall@5 of 88\% and an SR of 80\%, outperforming the best baseline by 7 percentage points in both metrics.


The average query-time inference latency, including the shared LLM call, was 2.0 s, compared with 21.3 s for end-to-end inference using GPT-4o.
The VRAM usage was 7.23 GB during querying and up to 7.40 GB during image embedding.
 
\begin{figure*}[t]
    \centering
    
    \includegraphics[clip,width=0.92\linewidth]{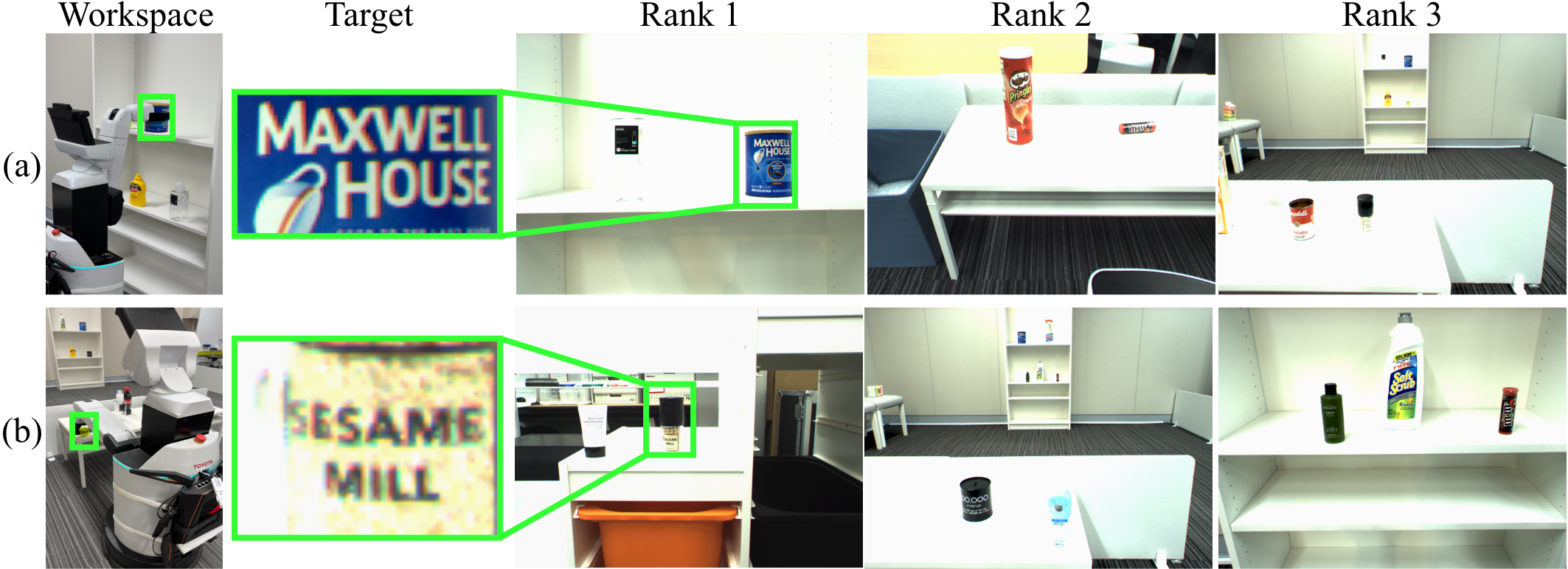}
    \caption{\textbf{Qualitative results for hardware experiments.}
    For the samples, $\bm{x}_{\text{txt}}$ values were as follows: (a) ``Pass me the Maxwell can.''; (b) ``Hand me the SESAME MILL.'' Each example shows the target object and the top-3 retrieved images. The green box indicates the target object.}
    \label{fig:hardware_success}
    \vspace{-2mm}
    
\end{figure*}

\textbf{Qualitative Results.}
Fig.~\ref{fig:hardware_success} depicts successful examples from the hardware experiments.
In Fig.~\ref{fig:hardware_success} (a), $\bm{x}_\text{txt}$ given to the robot was ``Pass me the Maxwell can.''
STARE correctly retrieved the target image at rank 1 using the partial product name ``Maxwell.''
Although the other retrieved images did not contain the scene text ``Maxwell,'' they still depicted a can, which aligned with the instruction.
In Fig.~\ref{fig:hardware_success} (b), where $\bm{x}_\text{txt}$ was ``Hand me the SESAME MILL,'' the model successfully retrieved the correct image, which included the scene text ``SESAME,'' at rank 1.
These results indicate that the proposed method effectively captures the intended target objects.
Following the retrieval of the target image, the DSR successfully grasped the target object.

%

%
\section{Conclusion}
{
We study STMR, the task of retrieving images with or without scene text from free-form instructions for robot manipulation or navigation.
STARE combines scene text-aware narrative representations with named-entity matching.
We constructed GoGetIt and TextCaps-test to evaluate this task.
STARE outperformed the evaluated baselines and achieved 80\% task success in zero-shot hardware experiments.

}

\subsubsection*{Acknowledgements.}
This work was partially supported by JST CREST and JSPS Fellows Grant Number JP25KJ2068.

\clearpage
\bibliographystyle{splncs04}
\bibliography{main}

\clearpage
\appendix
\setcounter{figure}{0}
\setcounter{table}{0}
\setcounter{equation}{0}
\renewcommand{\thefigure}{A\arabic{figure}}
\renewcommand{\thetable}{A\arabic{table}}
\renewcommand{\theequation}{A\arabic{equation}}
\renewcommand{\theHfigure}{A\arabic{figure}}
\renewcommand{\theHtable}{A\arabic{table}}
\renewcommand{\theHequation}{A\arabic{equation}}

\begin{center}
  {\Large\bfseries Supplementary Material}
\end{center}
\vspace{1em}

\section{Additional Related Works}
\label{sec:additional_related_works}

\subsection{Foundation Models for Robotics}

Recent studies have shown that remarkable generalization across manipulation and navigation tasks can be achieved by leveraging large-scale foundation models, such as large language models and vision-language models~\cite{brohan2023saycan, Driess2023PaLMEAE, brohan2022rt}. 
However, these approaches rely on a single foundation model, which can limit multi-level feature extraction. 
In contrast, STARE harnesses multiple foundation models to achieve a more comprehensive representation of the environment, tasks, and underlying semantic attributes.

\subsection{Multimodal Retrieval}

Numerous studies \cite{Yenamandra2023HomeRobotOM, robocup} have assessed the performance of domestic service robots in mobile manipulation tasks based on natural language instructions. 
Despite rapid advances in this area, there remains a critical gap in leveraging scene text (e.g., labels, signage) that inherently contains disambiguating information. 
STARE addresses this limitation by proposing a novel multimodal framework that explicitly incorporates scene text, enabling improved object retrieval accuracy in real-world environments.

\subsection{Scene Text}

Scene text retrieval methods often integrate OCR-extracted textual and positional information with visual features to improve alignment with query texts and achieve higher retrieval accuracy.
In contrast to scene text-based approaches, STARE integrates scene text and its position into a narrative representation that associates each text instance with the corresponding object and its attributes.
Furthermore, STARE introduces an auxiliary similarity function that efficiently models the complex relationships between scene texts and the referenced objects.

\subsection{Benchmarks}

Several standard benchmarks have been proposed in the fields of multimodal retrieval and tasks involving scene text.
Among the former, COCO~\cite{Lin2014MicrosoftCC} and Flickr30K~\cite{flickr30k} are originally image captioning benchmarks that have been widely used for multimodal retrieval tasks.
REVERIE~\cite{qi2020reverie} and GOAT-Bench~\cite{khanna2024goatbench} are designed for the object-goal navigation task, while LTRRIE~\cite{multirankit} focuses on multimodal retrieval by ranking images.
Benchmarks for scene text span various vision-language tasks.
For example, ST-VQA \cite{stvqa} is commonly used for scene text-based VQA, while TextCaps \cite{textcaps} and COCO-Text-Captioned \cite{stacmr} are suitable for scene text-aware image captioning.
Visual Genome~\cite{Krishna2017} includes images with scene texts and dense annotations, supporting vision-language grounding and reasoning tasks.
Finally, RefText~\cite{stan} is a benchmark for referring expression comprehension that provides image--object pairs and referring expressions.
Some of these expressions use scene text as a contextual cue.
Among the many existing benchmarks for multimodal retrieval, most are not suitable for our task because they lack scene text.
The benchmarks that include scene texts are typically not designed for multimodal retrieval, are unsuited to robotic tasks, or include overly simple instructions.
These limitations make them insufficient for evaluating real-world instruction-based retrieval tasks for robots, which require fine-grained reasoning over scene text and visual content.

\section{Task Details}
\label{sec:task_details}

\begin{figure}[t]
    \centering
    \includegraphics[clip,width=0.7\linewidth]{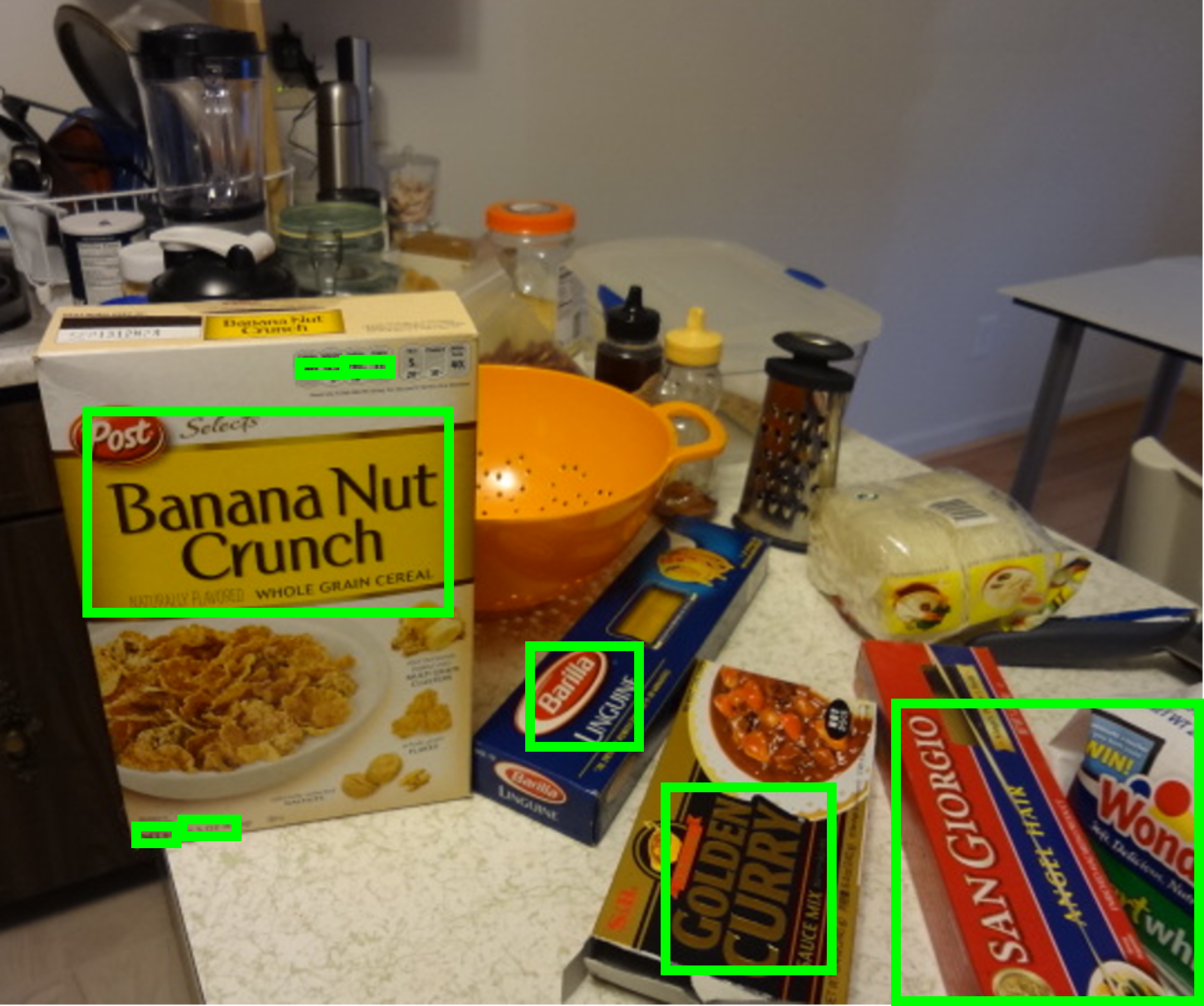}
    \caption{\textbf{An example of an input image} in the STMR task. The scene texts in the image are enclosed in green boxes.}
    \label{fig:task}
    
\end{figure}

In this study, we focus on the Scene Text-Aware Multimodal Retrieval (STMR) task, which involves retrieving images that contain the target object described in user-provided instructions.
Fig.~\ref{fig:task} shows an example of an input image in the STMR task.
For example, an instruction such as ``Please pass me the box of banana nut crunch placed in front of the orange bowl.'' is provided as input.
The model should retrieve images that align with the instruction and contain a box that has the scene text ``banana nut crunch.''

The STMR task is challenging because multimodal retrieval requires large-scale and fast inference, making it impractical to rely solely on MLLMs.
While existing methods \cite{clip, beit3} have achieved promising results in multimodal retrieval, their integration of scene text with other multimodal information is not always sufficient. 
This motivates explicitly modeling associations between scene text, objects, and their attributes.

The input to this task is a natural language instruction and a set of pre-collected images, and the output is a ranked list of the images based on their relevance to the instruction.
We use recall@$K$ ($K$=5,10,20) and task success rate as evaluation metrics.

\section{Method Details}
\label{sec:method_details}
\subsection{STVE}
\label{sec:method_details_stve}

The Scene Text-aware Visual Encoder (STVE) combines 
{features from image encoders, fine-grained features from overlapping patches, and scene text-aware narrative representations.}
Many everyday objects contain scene texts, making it essential to extract image features that effectively incorporate the scene texts present in the images.
However, scene text needs to be tied to the relevant objects rather than treated as independent OCR tokens.
OCR alone cannot reliably associate each detected scene-text instance with the correct object when an image contains multiple text instances.
Occlusions or unfavorable orientations could also result in inaccurate OCR-detected scene texts.
To address this, we introduce a scene text-aware narrative representation that associates OCR-detected scene text with the corresponding objects and their attributes.
This narrative representation can be generated using MLLMs.
However, relying solely on MLLMs may result in hallucination, where incorrect or nonexistent text is generated.
We mitigate this issue by providing the MLLM with the scene texts detected by a third-party OCR and visually marking their locations within the image through visual prompting.
This approach introduces a different challenge: the MLLM may mistakenly interpret the overlaid visual prompts as part of the actual scene texts, resulting in erroneous text extraction.
To address this limitation, the STVE module generates a narrative representation using CVP.

{The prompt used by the ``Image explanation embedding'' in Fig.~\ref{fig:model} of the main paper to obtain a narrative representation} from $\bm{x}_{\text{img}}^{(i)}$ was as follows:
\begin{nolinenumbers}
\begin{lstlisting}[aboveskip=8pt,belowskip=8pt,breaklines=true,breakindent=0pt, breakautoindent=false,frame=single,basicstyle=\ttfamily\footnotesize]
Describe the objects in this image in detail, including their positional relationships to surrounding objects. 
The positional labels in the image and the OCR-detected text pairs are as follows: {`, '.join(indexed ocr-detected text list)}. 
Each positional label, shown in white text on a black background, is located at the center of the OCR text detection area and above its top edge.
If multiple texts are associated with a single positional label, identify the correct text for that label. 
Do not explain using position labels, but use them only to understand the image. 
Output the description in English.
\end{lstlisting}
\end{nolinenumbers}

\section{Loss Function}
\label{sec:loss_function}

We used the Double Relaxed Contrastive loss~\cite{relaxformer}, replacing cosine similarity with the STRR score $s$, including the lexical matching term.
The loss used in this study is defined as follows: 
\begin{align}
\mathcal{L} &= \mathcal{L}_\text{P} + \gamma\mathcal{L}_\text{UP} + \beta\mathcal{L}_\text{N}\\
\mathcal{L}_\text{P}   &= \sum_j \left( 1 - s \left( \boldsymbol{x}_\text{txt}^{(j)}, \boldsymbol{x}_\text{img}^{(i)} \right) \right)^2 \\
\mathcal{L}_\text{UP}  &= \sum_{(i,j) \in U} \max \left( \alpha - s \left( \boldsymbol{x}_\text{txt}^{(j)}, \boldsymbol{x}_\text{img}^{(i)} \right), 0 \right)^2 \\
\mathcal{L}_\text{N}   &= \sum_{(i,j) \notin{U}} \max \left( s \left( \boldsymbol{x}_\text{txt}^{(j)}, \boldsymbol{x}_\text{img}^{(i)} \right), 0 \right)^2,
\end{align}
where $\gamma$ and $\beta$ weight the unlabeled-positive and negative losses, respectively; $U$ is the set of unlabeled-positive index pairs; and $\alpha$ is the margin for unlabeled-positive pairs.

\clearpage
\section{Annotation Details}
\label{sec:benchmark_annotation}
\setlength{\intextsep}{0pt}
\begin{wrapfigure}[22]{r}{0.45\textwidth}
    \centering
    \includegraphics[clip,width=\linewidth]{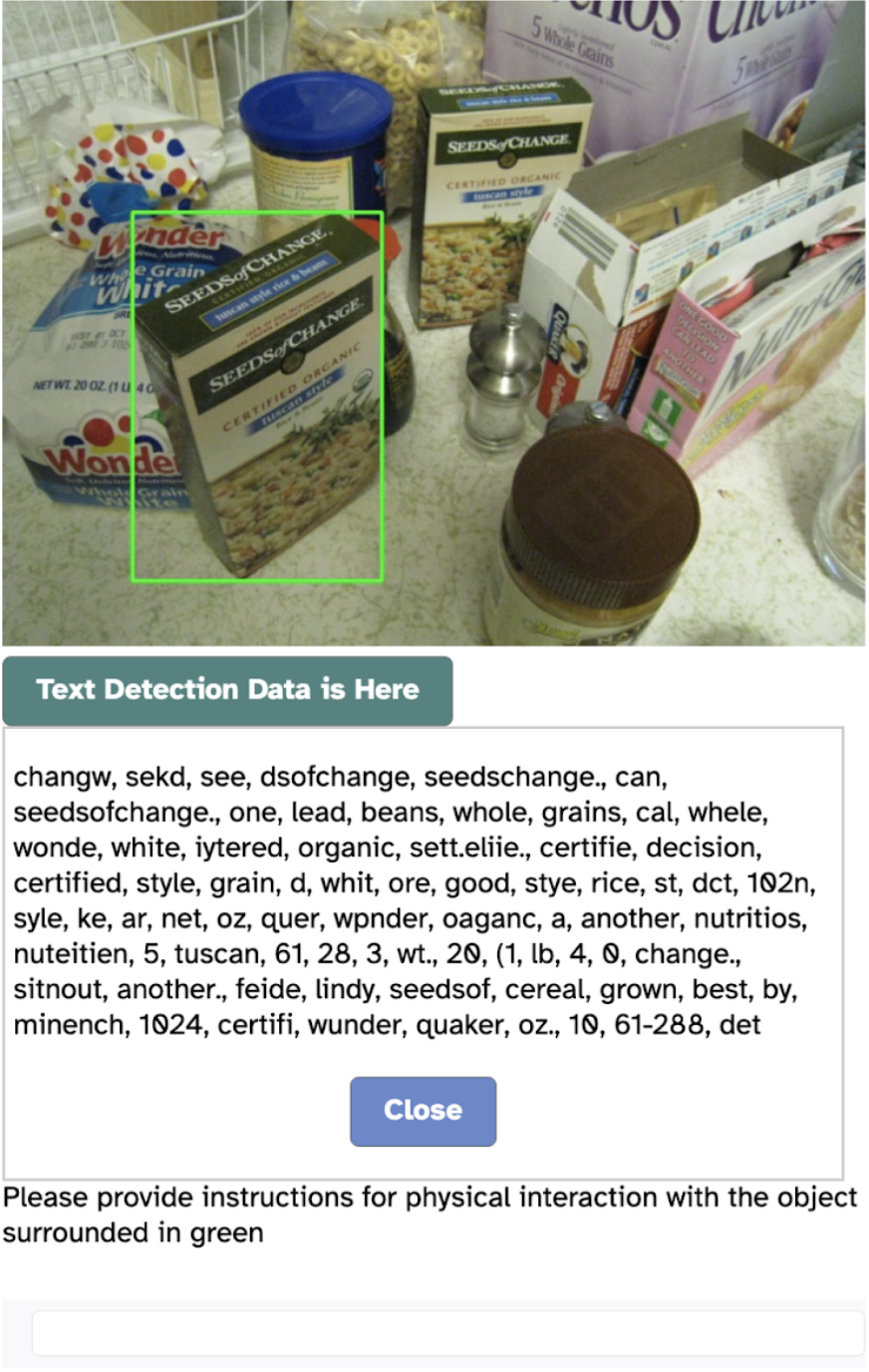}
    \caption{Annotation interface. Annotators were asked to give object manipulation or navigation instructions with scene text provided as optional context.}
    \label{fig:ui}
\end{wrapfigure}
The Instruction subset of the GoGetIt dataset and the TextCaps-test dataset were annotated with instructions.
Fig.~\ref{fig:ui} presents the annotation interface used to crowdsource the instructions. Annotators were shown images with the
target object enclosed in a bounding box and scene texts extracted from the image. They were then asked to input
full-sentence instructions for diverse object manipulation or navigation actions, optionally mentioning scene text.
An example of an instruction for the sample in the figure could be ``Pass me the seeds of change box in front of the bag of wonder bread.''
{\color{fix}To further ensure annotation quality, all test samples were subsequently reviewed by multiple experts against predefined quality-check criteria, requiring each instruction to be a full-sentence manipulation or navigation command consistent with the target object and scene text context.}

\par\vspace{1.5\baselineskip}
\section{Experimental Setup Details}

\subsection{Baseline Methods}
We selected CLIP, Long-CLIP, BLIP-2, and BEiT-3 because they are representative methods for zero-shot multimodal retrieval tasks~\cite{clip, longclip, blip2, beit3}.
Additionally, NLMap and RelaX-Former were selected for their use in related everyday-object identification~\cite{multirankit} and image-retrieval-based open-vocabulary fetch-and-carry tasks~\cite{relaxformer}, respectively.

\subsection{Hardware Experimental Setup}
\label{hardware_setup}

\begin{figure}[!t]
    \centering
    \includegraphics[clip,width=0.6\linewidth]{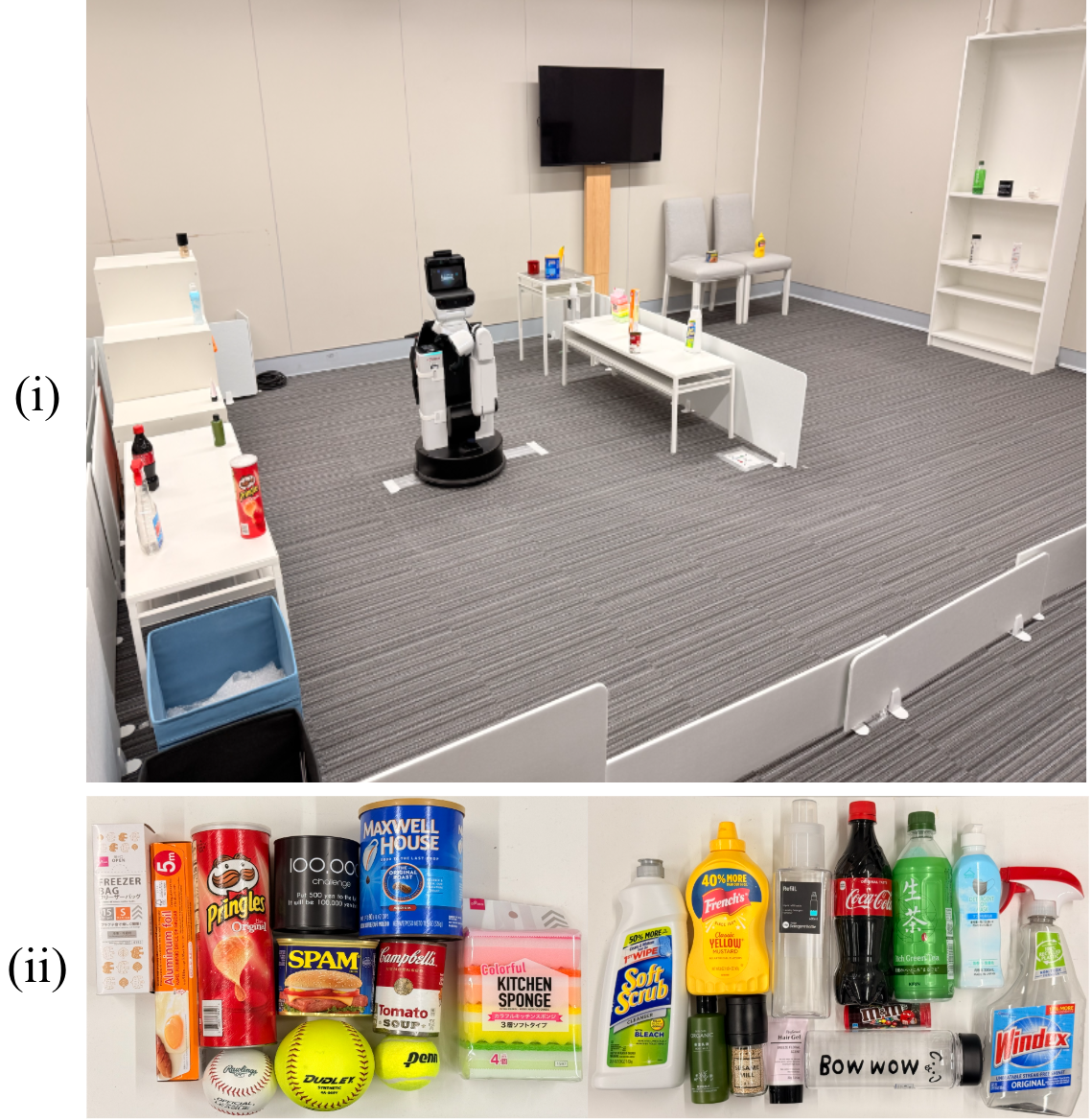}
    \caption{\textbf{Experimental settings.} (i) A domestic environment as specified in WRS2020RS~\cite{wrs2020}. (ii) Objects selected for the hardware experiments, consisting of YCB objects~\cite{calli2015ycb} and everyday objects with scene texts.}
    \label{fig:hardware_setting}
\end{figure}

Fig.~\ref{fig:hardware_setting} shows the hardware setup used in the real-world experiments.
The environment replicated the standardized layout of the World Robot Summit 2020 Partner Robot Challenge/Real Space (WRS2020RS~\cite{wrs2020}), a domestic setting designed for benchmark tidy-up tasks.
The environment measured $4.0 \times 6.0~\text{m}^2$ and contained nine fixed furniture items arranged according to the official layout.
Experiments were conducted using the Human Support Robot (HSR) developed by Toyota Motor Corporation.
The HSR has served as the standard platform for RoboCup@Home~\cite{I15} since 2017.
The object set contained 23 everyday objects: 10 from the YCB object set~\cite{calli2015ycb} and 13 additional household items with printed labels.
We included the latter to evaluate performance on objects with scene text.

\subsection{Evaluation Metric}
\label{sec:evaluation_metric}

%
We used recall@$K$ ($K$=5,10,20) as the evaluation metric, with recall@10 as the primary metric.
recall@$K$ is defined as follows:
\begin{align}
\text{recall@}K = \frac{1}{N_{\text{q}}} \sum_{j=1}^{N_{\text{q}}} \mathbf{1}\!\left[A_j \cap B_j^{(K)} \neq \varnothing\right],
\end{align}
Here, $N_\text{q}$ is the number of instructions, $A_j$ is the set of target images for instruction $j$, and $B_j^{(K)}$ is its top-$K$ retrieved image set. The indicator is 1 if at least one target image is retrieved, and 0 otherwise. Scores are computed separately for each benchmark.

\section{Additional Results}

\subsection{Additional Quantitative Results}
\label{sec:additional_quantitative_results}

\begin{table}[!t]
\setlength{\tabcolsep}{1.8pt}
\centering
\caption{{\color{fix}\textbf{Additional quantitative results} between STARE and baseline methods on the RefText subset of the GoGetIt benchmark, the Instruction subset of the GoGetIt benchmark, TextCaps-test benchmark, and LTRRIE dataset. The recall@20 scores are shown, with the best score for each benchmark in bold. `*' and `$\dagger$' denote reproduced results and fine-tuned models, respectively.}}
\label{tab:results_appendix}
\small
{
\begin{tabular}{lccccc}
\toprule
\hspace{-2pt}{[\%]}
& {GoGetIt}  
& {GoGetIt}  
& {TextCaps}  
& \multirow{2}{*}{LTRRIE} 
\\ 
Method
& {(RefText)}
& {(Instruction)}
& {-test}
&
\\

\midrule 
CLIP \cite{clip}
& {82.3} 
& {92.1}
& {91.5}
& {84.6}
\\

CLIP $\dagger$ 
& {83.3\scriptsize~$\pm$~{0.2}} 
& {90.9\scriptsize~$\pm$~{0.3}}
& {93.5\scriptsize~$\pm$~{0.3}} 
& {84.6\scriptsize~$\pm$~{0.3}}
\\

{NLMap* \cite{nlmap}}
& {70.5} 
& {86.3} 
& {85.8}
& {78.8} 
\\
{BEiT-3 \cite{beit3}} 
& {76.6} 
& {89.9}
& {91.3}
& {88.3} 
\\
{BLIP-2 \cite{blip2}} 
& {-} 
& {-} 
& {94.8}
& {91.0} 
\\

Long-CLIP \cite{longclip}
& {-} 
& {-} 
& {94.8}
& {91.0} 
\\

Long-CLIP $\dagger$ 
& {-} 
& {-} 
& {95.9\scriptsize~$\pm$~{0.1}}
& {94.1\scriptsize~$\pm$~{0.1}}
\\
RelaX-Former \cite{relaxformer}
& {-} 
& {-} 
& {84.9\scriptsize~$\pm$~{0.9}}
& {92.3\scriptsize~$\pm$~{0.5}}
\\

{{\color{fix}ReMEmbR* \cite{anwar2025remembr}}}
& {\color{fix}57.9} 
& {\color{fix}68.1} 
& {\color{fix}86.3}
& {\color{fix}55.9} 
\\
{{\color{fix}Meta CLIP 2 \cite{chuang2025metaclip2}}}
& {\color{fix}85.4}
& {\color{fix}91.3}
& {\color{fix}95.8}
& {\color{fix}88.7}
\\

{{\color{fix}ViSTA* $\dagger$ \cite{vista}}}
& {\color{fix}89.1\scriptsize~$\pm$~{0.4}}
& {\color{fix}90.5\scriptsize~$\pm$~{0.4}}
& {\color{fix}93.5\scriptsize~$\pm$~{0.3}}
& {\color{fix}91.4\scriptsize~$\pm$~{0.5}}
\\

STARE (ours)
& \boldsymbol{{96.9}}\scriptsize~$\pm$~{0.3}
& \boldsymbol{{95.6}}\scriptsize~$\pm$~{0.2}
& \boldsymbol{{98.3}}\scriptsize~$\pm$~{0.4}
& \boldsymbol{{95.4}}\scriptsize~$\pm$~{0.9}
\\
\bottomrule
\end{tabular}
}
\end{table}

{
Table~\ref{tab:results_appendix} presents additional quantitative results between STARE and the baseline methods using recall@20.
The scores for CLIP (frozen), Long-CLIP (frozen), BLIP-2, BEiT-3, Meta CLIP 2, ReMEmbR, and NLMap are reported from a single trial because their scores were consistent across repeated runs.
The mean and standard deviation from five trials are reported for the other methods.
The highest score for each benchmark is highlighted in bold.
The scores for Long-CLIP, BLIP-2, and RelaX-Former are not reported for the GoGetIt benchmark due to data leakage between their training data and the test set of the GoGetIt benchmark.
Table~\ref{tab:results_appendix} shows that STARE consistently outperformed all baseline methods across the RefText subset of the GoGetIt benchmark, the Instruction subset of the GoGetIt benchmark, the TextCaps-test benchmark, and the LTRRIE dataset, achieving recall@20 scores of 96.9\%, 95.6\%, 98.3\%, and 95.4\%, respectively.
}

\subsection{Additional Qualitative Results}
\label{sec:additional_qualitative_results}

\begin{figure}[!t]
    \centering
    \includegraphics[clip,width=\linewidth]{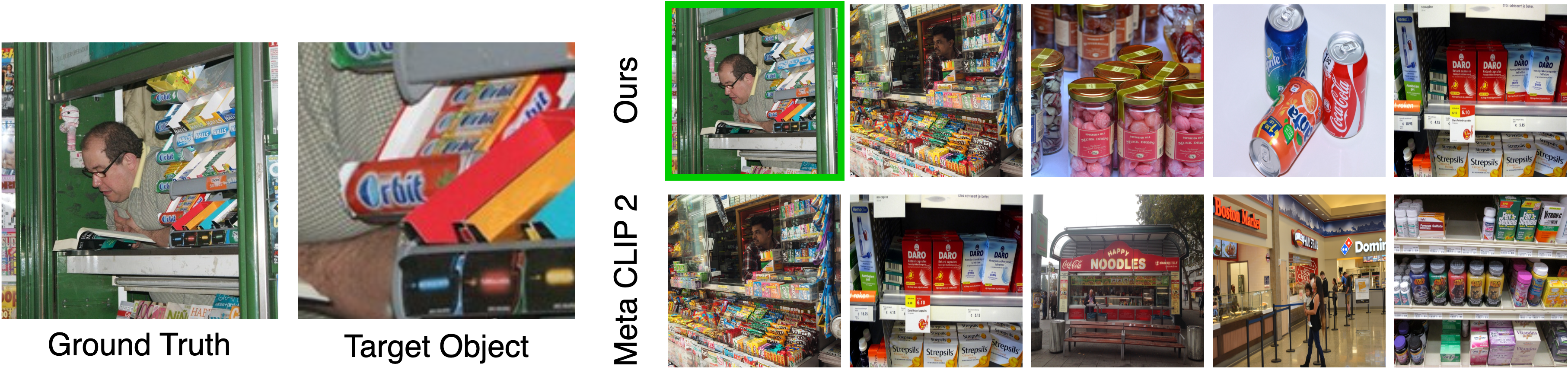}
    \caption{\textbf{Qualitative results} of the proposed method and a baseline method on the TextCaps-test benchmark. For the sample, $\bm{x}_{\text{txt}}$ was ``Buy the red Orbit candy at the kiosk.''}
    \label{fig:success-1}
\end{figure}

Fig.~\ref{fig:success-1} presents additional qualitative results comparing the proposed method and a baseline method on a sample from the TextCaps-test benchmark.
The figure shows the target image and the top-3 images retrieved by each method.
If retrieved, the target image is highlighted with a green border.
In this example, $\bm{x}_{\text{txt}}$ was ``Buy the red Orbit candy at the kiosk.''
The baseline method ranked images that included a kiosk or the color red within the top-3, but none of these images contained a red Orbit candy.
In contrast, STARE successfully ranked the target image, which contains red Orbit candies, as the top result.

\begin{figure}[!t]
    \centering
    \includegraphics[clip,width=0.8\linewidth]{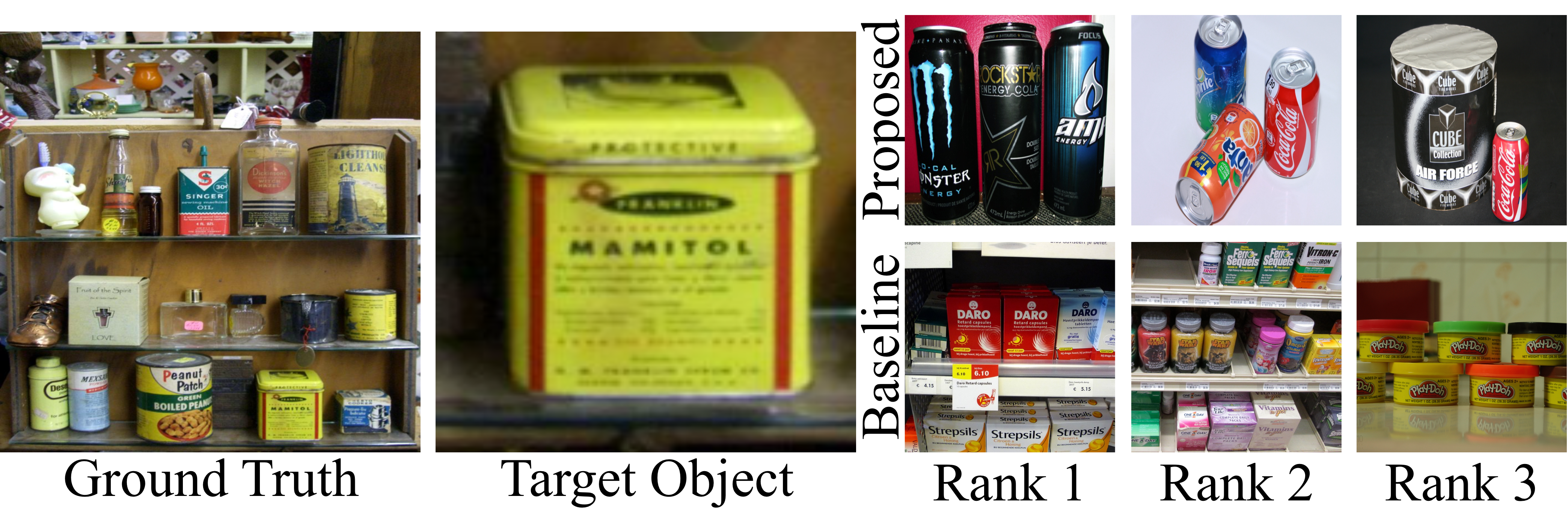}
    \caption{A sample from the TextCaps-test benchmark, in which the proposed method \textbf{failed to rank} the target image within the top-3. The given $\bm{x}_{\text{txt}}$ was ``The square yellow mamitol can with red lines.''
    }
    
    \label{fig:failed}
\end{figure}

Fig.~\ref{fig:failed} shows an example in which the proposed method failed to rank the target image within the top-3.
For this sample, $\bm{x}_{\text{txt}}$ was: ``The square yellow mamitol can with red lines.''
The figure presents the corresponding target image and the target object within the image, along with the top-3 images retrieved by the proposed method and a baseline method.
The target object is a square yellow tin container labeled with the scene text ``MAMITOL.'' Red lines run vertically along the left and right edges of its front.
The proposed method ranked an image containing a red can as rank 1, a row of red telephone boxes as rank 2, and cans lined up against a red background as rank 3.
The target image was not ranked in the top-3.
One likely cause is that the model could not use the unfamiliar scene text ``MAMITOL'' to extract useful multimodal features.

\subsection{Additional Ablation Study}
\label{sec:additional_ablation_study}

Ablation studies on the Scene Text ReRanker (STRR) module and Crosslingual Visual Prompt (CVP) were also performed using the RefText subset of the GoGetIt benchmark and LTRRIE~\cite{multirankit} dataset.
Additionally, we performed ablation on the patchification in the Scene Text-aware Visual Encoder (STVE) module.

\begin{table*}[t]
\centering
\caption{
\textbf{Results of ablation study}. The average and standard deviation from five trials are reported. Bold indicates the highest value. (a)-(l) correspond to model variants for the ablation study. K, AN, E, G, BB, QE denote Katakana, Arabic numerals, English, Greek, bounding box, and query expansion, respectively. 
}
\label{tab:ablation2}
\resizebox{\textwidth}{!}
{
\small
\begin{tabular}{lcccccccccccc}
\toprule
\hspace{-2pt}{[\%]}
& \multirow{2}{*}{STRR} 
& \multirow{2}{*}{CVP} 
& \multirow{2}{*}{QE} 
& \multirow{2}{*}{Patchification} 
& \multicolumn{2}{c}{GoGetIt(RefText)}  
& \multicolumn{2}{c}{GoGetIt(Instruction)}  
& \multicolumn{2}{c}{TextCaps-test}  
& \multicolumn{2}{c}{LTRRIE} 
\\ 
\cmidrule(lr){6-7}
\cmidrule(lr){8-9}
\cmidrule(lr){10-11}
\cmidrule(lr){12-13}
Model
& 
&
&
&
& R@5 $\uparrow$  
& R@10 $\uparrow$ 
& R@5 $\uparrow$  
& R@10 $\uparrow$
& R@5 $\uparrow$  
& R@10 $\uparrow$   
& R@5 $\uparrow$  
& R@10 $\uparrow$

\\
\hline
(a)  
& $\boldsymbol{T}_{\text{NE}}$
& K
& \checkmark
& overlap
& \textbf{90.2}\scriptsize~$\pm$~0.4
& \textbf{94.4}\scriptsize~$\pm$~0.4
& \textbf{89.8}\scriptsize~$\pm$~0.6
& \textbf{92.8}\scriptsize~$\pm$~0.6
& 92.4\scriptsize~$\pm$~1.3
& \textbf{95.4}\scriptsize~$\pm$~0.6
& \textbf{72.5}\scriptsize~$\pm$~0.8
& \textbf{86.6}\scriptsize~$\pm$~0.9

\\
\hline

(b) 
& - 
& K
& \checkmark
& overlap
& 76.4\scriptsize~$\pm$~0.8
& 85.0\scriptsize~$\pm$~0.7
& 78.2\scriptsize~$\pm$~1.2
& 85.9\scriptsize~$\pm$~0.5
& 83.6\scriptsize~$\pm$~1.1
& 89.5\scriptsize~$\pm$~1.9
& 71.1\scriptsize~$\pm$~1.0
& 85.1\scriptsize~$\pm$~0.6
\\

(c)  
& $\boldsymbol{T}_{\text{txt}}$
& K
& \checkmark
& overlap
& 81.4\scriptsize~$\pm$~0.8
& 89.2\scriptsize~$\pm$~0.6
& 79.7\scriptsize~$\pm$~1.1
& 87.7\scriptsize~$\pm$~1.1
& 82.1\scriptsize~$\pm$~0.6
& 89.4\scriptsize~$\pm$~1.0
& 70.6\scriptsize~$\pm$~1.3
& 85.1\scriptsize~$\pm$~0.8
\\

\hline

(d)  
& $\boldsymbol{T}_{\text{NE}}$
& AN
& \checkmark
& overlap
& \textbf{90.2}\scriptsize~$\pm$~0.6
& 94.3\scriptsize~$\pm$~0.2
& 88.4\scriptsize~$\pm$~0.5
& 92.3\scriptsize~$\pm$~0.8
& 92.3\scriptsize~$\pm$~1.0
& 95.3\scriptsize~$\pm$~0.4
& 71.3\scriptsize~$\pm$~1.0
& 85.9\scriptsize~$\pm$~0.3
\\

(e)
& $\boldsymbol{T}_{\text{NE}}$
& E
& \checkmark
& overlap
& 90.1\scriptsize~$\pm$~0.9
& 94.3\scriptsize~$\pm$~0.4
& 89.3\scriptsize~$\pm$~0.5
& 92.2\scriptsize~$\pm$~0.6
& 91.3\scriptsize~$\pm$~0.6
& 95.3\scriptsize~$\pm$~0.6
& 71.6\scriptsize~$\pm$~0.9
& 85.7\scriptsize~$\pm$~1.3
\\

(f)
& $\boldsymbol{T}_{\text{NE}}$
& G
& \checkmark
& overlap
& 90.1\scriptsize~$\pm$~0.8
& 94.3\scriptsize~$\pm$~0.3
& 88.9\scriptsize~$\pm$~0.4
& 92.1\scriptsize~$\pm$~0.5
& 91.9\scriptsize~$\pm$~0.9
& 95.3\scriptsize~$\pm$~0.5
& 71.2\scriptsize~$\pm$~1.0
& 86.0\scriptsize~$\pm$~0.5
\\

(g)  
& $\boldsymbol{T}_{\text{NE}}$
& BB
& \checkmark
& overlap
& 90.0\scriptsize~$\pm$~0.2
& 94.2\scriptsize~$\pm$~0.2
& 89.1\scriptsize~$\pm$~0.8
& 92.6\scriptsize~$\pm$~0.5
& 91.9\scriptsize~$\pm$~1.1
& 95.2\scriptsize~$\pm$~0.1
& 70.9\scriptsize~$\pm$~2.0
& 85.5\scriptsize~$\pm$~1.4
\\

\hline

(h) 
& $\boldsymbol{T}_{\text{NE}}$
& -
& \checkmark
& overlap
& 87.0\scriptsize~$\pm$~0.4
& 92.1\scriptsize~$\pm$~0.5
& 87.3\scriptsize~$\pm$~0.6
& 91.9\scriptsize~$\pm$~0.5
& 89.5\scriptsize~$\pm$~1.0
& 93.7\scriptsize~$\pm$~0.5
& 67.2\scriptsize~$\pm$~1.5
& 82.3\scriptsize~$\pm$~0.7
\\

(i) 
& -
& -
& \checkmark
& overlap
& 69.3\scriptsize~$\pm$~1.1
& 79.7\scriptsize~$\pm$~0.5
& 73.6\scriptsize~$\pm$~0.8
& 82.3\scriptsize~$\pm$~1.2
& 77.6\scriptsize~$\pm$~0.9
& 84.8\scriptsize~$\pm$~1.3
& 67.4\scriptsize~$\pm$~0.4
& 81.9\scriptsize~$\pm$~0.6
\\
\hline

(j) 
& $\boldsymbol{T}_{\text{NE}}$
& K
& -
& overlap
& 89.6\scriptsize~$\pm$~0.7
& 93.9\scriptsize~$\pm$~0.3
& 88.8\scriptsize~$\pm$~0.3
& 92.6\scriptsize~$\pm$~0.8
& 91.9\scriptsize~$\pm$~0.6
& 95.3\scriptsize~$\pm$~0.7
& 68.8\scriptsize~$\pm$~0.9
& 83.5\scriptsize~$\pm$~0.8
\\
\hline

(k) 
& $\boldsymbol{T}_{\text{NE}}$
& K
& \checkmark
& -
& 89.9\scriptsize~$\pm$~0.7
& 94.2\scriptsize~$\pm$~0.6
& 89.5\scriptsize~$\pm$~0.7
& 92.2\scriptsize~$\pm$~0.5
& \textbf{92.5}\scriptsize~$\pm$~0.5
& 95.3\scriptsize~$\pm$~0.1
& 71.3\scriptsize~$\pm$~1.0
& 84.8\scriptsize~$\pm$~0.6
\\

(l) 
& $\boldsymbol{T}_{\text{NE}}$
& K
& \checkmark
& standard
& 88.4\scriptsize~$\pm$~1.2
& 93.4\scriptsize~$\pm$~0.7
& 88.8\scriptsize~$\pm$~0.6
& 92.4\scriptsize~$\pm$~0.4
& 90.7\scriptsize~$\pm$~0.5
& 94.9\scriptsize~$\pm$~0.6
& 68.3\scriptsize~$\pm$~0.7
& 83.7\scriptsize~$\pm$~0.6
\\

\bottomrule
\end{tabular}
}
\end{table*}

\paragraph{MFIE Ablation.}
For the MFIE ablation, Model (j) removes {query expansion}.
Table~\ref{tab:ablation2} shows that the recall@10 score for Model (j) on the Instruction subset of the GoGetIt benchmark was 92.6\%, which was 0.2 points lower than that of Model (a).
This result suggests that using {query expansion} helps align named entities with visual characteristics by describing attributes of the entities.

\paragraph{Patchification ablation.}
To determine the contribution of using patchification, we removed this functionality in Model (k).
Within models using patchification, the effects of overlapping patches were examined by replacing overlapped patchification with standard patchification in Model (l).
Table~\ref{tab:ablation2} indicates that the recall@10 scores on the Instruction subset of the GoGetIt benchmark were 92.2\% for Model (k) and 92.4\% for Model (l).
These values are 0.6 and 0.4 points lower, respectively, than the score achieved by Model (a).
These results demonstrate that overlapped patchification improves the model’s ability to capture detailed visual features, particularly those located at the patch boundaries.

\paragraph{MLLM Backbone Ablation.}
We used GPT-4o~\cite{gpt_4o} for narrative generation, query expansion, and named-entity extraction in the main experiments.
We replaced it in all three roles with Gemma-4-E4B-it~\cite{gemma4} or Qwen3.5-9B~\cite{qwen3_5} and retrained STARE with the resulting representations.

\begin{table}[t]
\centering
\caption{
\textbf{Recall@10 with different MLLMs for narrative generation, query expansion, and named-entity extraction.} The average and standard deviation from five trials are reported. Bold indicates the highest value.
}
\label{tab:ablation_mllm}
\small
\begin{tabular}{lcc}
\toprule
Backbone MLLM & LTRRIE & TextCaps-test \\
& R@10 $\uparrow$ [\%] & R@10 $\uparrow$ [\%] \\
\midrule
GPT-4o~\cite{gpt_4o} & \textbf{86.6}\scriptsize~$\pm$~0.9 & \textbf{95.4}\scriptsize~$\pm$~0.6 \\
Gemma-4-E4B-it~\cite{gemma4} & 83.4\scriptsize~$\pm$~1.3 & 94.8\scriptsize~$\pm$~1.6 \\
Qwen3.5-9B~\cite{qwen3_5} & 84.9\scriptsize~$\pm$~2.2 & 93.8\scriptsize~$\pm$~1.2 \\
\bottomrule
\end{tabular}
\end{table}

As shown in Table~\ref{tab:ablation_mllm}, both alternatives remained within 3.2 recall@10 points of GPT-4o across the LTRRIE dataset and the TextCaps-test benchmark, indicating limited sensitivity to the choice of MLLM.

\section{Error Analysis}
\label{sec:error_analysis}

To investigate the limitations of the proposed method, we analyzed cases in which STARE did not perform as expected. 
We defined a failure as an instruction for which no target image was retrieved within the top 10.
There were a total of 411 failure cases, with 81, 265, 15, and 50 samples from the RefText subset of the GoGetIt benchmark, the Instruction subset of the GoGetIt benchmark, the TextCaps-test benchmark, and the LTRRIE~\cite{multirankit} dataset, respectively.
We selected the 100 worst-ranked failure cases for analysis. Table~\ref{tab:error} categorizes their failure modes. The failure modes could be grouped into the following seven categories:

\begin{table}[t]
\captionsetup{width=0.45\textwidth}
\centering
\caption{Number of failure cases in each category.}
\label{tab:error}
\begin{tabular}{lc}
\toprule
Error mode
& \#Error

\\ \hline
Attribute comprehension error
& 25

\\
Annotation error
& 21

\\Named entity extraction error
& 15

\\OCR detection error
& 15

\\Ambiguous instruction
& 14

\\
Numerical comprehension error
& 6

\\
Scene text comprehension error
& 4

\\ \hline
Total
& 100

\\ \bottomrule
\end{tabular}
\end{table}

\paragraph{Attribute comprehension error.} 
This category covers cases in which the model fails to correctly understand or identify the attributes of objects, such as color, category, shape and spatial relationships.

\paragraph{Annotation error.} 
This category covers inaccurate annotations: the instruction refers to an object or scene text absent from the target image.

\paragraph{Named entity extraction error.} 
This refers to cases in which the model fails to accurately extract named entities from the instruction.

\paragraph{OCR detection error.} 
This involves cases in which the OCR system fails to detect or correctly recognize scene texts, leading to incomplete or inaccurate text extraction.

\paragraph{Ambiguous instruction.} 
This category refers to cases in which multiple images in the retrieval set could potentially match the given instruction, making it unclear which image is the intended target.

\paragraph{Numerical comprehension error.} 
This category includes cases in which the model fails to correctly comprehend numerical information.

\paragraph{Scene text comprehension error.}
This category covers cases in which OCR detects the scene text, but the model fails to interpret or use it for retrieval and reranking.

Table~\ref{tab:error} indicates that the main bottleneck lies in attribute comprehension errors.
A possible reason is that images in the dataset differ substantially in appearance.
Such differences can allow retrieval without fine-grained attributes such as color or spatial relationships.
As a result, the model may lack the ability to capture such detailed attribute information.
A possible solution is to perform data augmentation by modifying the color and spatial relationships of target objects in the instructions and target images.

\section{Additional Hardware Comparison}
\label{sec:additional_hardware_comparison}
\begin{table}[!t]
\captionsetup{width=0.75\textwidth, font=small}
\centering
\scriptsize
\caption{
    \textbf{Quantitative hardware comparison} between STARE and GPT-4o.
The scores are the results from 100 experimental runs.
}
{
\begin{tabular}{lccc}
\toprule
Method & R@5 $\uparrow$ [\%] & SR $\uparrow$ [\%] & Inference time $\downarrow$ [s] \\
\midrule
STARE (ours) & 88 & 80 & 2.0 \\
GPT-4o & 45 & 43 & 21.3 \\
\bottomrule
\end{tabular}
}
\label{tab:mllm_comparison}
\end{table}

Table~\ref{tab:mllm_comparison} compares STARE with GPT-4o in the hardware setting.
STARE achieved a recall@5 of 88\% and an SR of 80\%, outperforming GPT-4o by 43 and 37 points, respectively.
In addition, STARE required only 2.0 seconds for inference, compared with 21.3 seconds for GPT-4o, demonstrating its practical efficiency for real-world robotic deployment.
Query expansion and named-entity extraction are performed jointly in a single LLM call.
This call accounts for approximately 1.6 s of the 2.0 s query-time latency.

\section{Limitations}

\label{sec:limitations}
{
Although STARE achieved promising results, several limitations remain in the reranking process.
First, the model struggles to handle numerical expressions such as price tags and ordinal numbers.
Although numerical reasoning remains challenging even for modern MLLMs, possible solutions include prompting chain-of-thought reasoning to process numerical values and their associated meanings, as well as incorporating training on relevant datasets.
In addition, LLMs exhibit limited robustness when handling unknown named entities, which can result in hallucinations when generating {query expansion} of target objects in the MFIE module.
A possible solution is 
to complement the LLM’s internal world knowledge with external information.
}

\end{document}